%% file: main.tex
\documentclass[10pt,twocolumn,letterpaper]{article}
\PassOptionsToPackage{numbers, compress}{natbib}

\usepackage{cvpr}              % To produce the CAMERA-READY version
\usepackage{float}

\definecolor{mygray}{gray}{.92}

\usepackage{changes}
\usepackage{csquotes}

\input{preamble}

\definecolor{cvprblue}{rgb}{0.21,0.49,0.74}
\usepackage[pagebackref,breaklinks,colorlinks,allcolors=cvprblue]{hyperref}
\usepackage{cancel}
\usepackage{listings}

\definecolor{codegreen}{rgb}{0,0.6,0}
\definecolor{codegray}{rgb}{0.5,0.5,0.5}
\definecolor{codepurple}{rgb}{0.58,0,0.82}
\definecolor{backcolour}{rgb}{0.95,0.95,0.92}
\lstdefinestyle{mystyle}{
    backgroundcolor=\color{backcolour},   
    commentstyle=\color{codegreen},
    keywordstyle=\color{magenta},
    numberstyle=\tiny\color{codegray},
    stringstyle=\color{codepurple},
    basicstyle=\ttfamily\footnotesize,
    breakatwhitespace=false,         
    breaklines=true,                 
    captionpos=b,                    
    keepspaces=true,                 
    numbers=left,                    
    numbersep=5pt,                  
    showspaces=false,                
    showstringspaces=false,
    showtabs=false,                  
    tabsize=2,
    inputencoding=utf8,
    escapeinside={(*@}{@*)}, 
}
\def\paperID{} % *** Enter the Paper ID here
\def\confName{CVPR}
\def\confYear{2026}

\vspace{-4mm}

\title{SpatialVID: A Large-Scale Video Dataset with Spatial Annotations}

\author{
Jiahao Wang$^{1\dagger}$~~~
Yufeng Yuan$^{1\dagger}$~~~
Rujie Zheng$^{1\dagger}$~~~
Youtian Lin$^{1}$~~~
Jian Gao$^{1}$~~~\\
Lin-Zhuo Chen$^{1}$~~~
Yajie Bao$^{1}$~~~
Yi Zhang$^{1}$~~~
Chang Zeng$^{1}$~~~
Yanxi Zhou$^{1}$~~~\\
Xiao-Xiao Long$^{1}$~~~
Hao Zhu$^{1}$~~~
Zhaoxiang Zhang$^{2}$~~~
Xun Cao$^{1}$~~~
Yao Yao$^{1\ddagger}$~~~
\\
\small{
$^{1}$Nanjing University \ 
$^{2}$Institute of Automation, Chinese Academy of Sciences \
$^{\dagger}$Equal contribution \
$^{\ddagger}$Corresponding author
}
}
\newcommand{\DATAname}{SpatialVID}
\begin{document}

% \vspace{-6mm}

\twocolumn[{%
\maketitle
% \vspace{-2mm}
% \vspace{-5mm}

\thispagestyle{empty}
\begin{center}
\centering
\includegraphics[width=.95\textwidth]{figure/overview.pdf}
% \vspace{-5mm}
\captionof{figure}{We introduce \DATAname, a 
  large-scale video dataset with explicit spatial annotations including camera poses, depth maps, structured captions and serialized motion instructions.
  The dataset consists of 7,089 hours of real-world dynamic scenes. An exemplar of our \DATAname~is shown on the right.
}%
\label{fig:teaser}%
\end{center}%
}]

\input{sec/0_abstract}
\input{sec/1_introduction}

\input{sec/2_relatedwork}

\input{sec/3_curation}
\input{sec/4_analysis}

\input{sec/5_task1}

\input{sec/5_task2}

\input{sec/5_task3}

\input{sec/7_conclusion}

{
    \small
    
    \bibliographystyle{ieeenat_fullname}
    \bibliography{ref}
}
\input{sec/suppl}

% \bibliography{ref}

% WARNING: do not forget to delete the supplementary pages from your submission 
\end{document}

%% file: preamble.tex
\newcommand{\figref}[1]{Fig.~\ref{#1}}
\newcommand{\tabref}[1]{Tab.~\ref{#1}}
\newcommand{\secref}[1]{Sec.~\ref{#1}}

\usepackage{overpic}
\usepackage{tikz}
\usepackage[utf8]{inputenc} % allow utf-8 input
\usepackage{changes}
\usepackage{csquotes}
\usepackage{tikz}
\usepackage[utf8]{inputenc} % allow utf-8 input
\usepackage[T1]{fontenc}    % use 8-bit T1 fonts
\usepackage{url}            % simple URL typesetting
\usepackage{booktabs}       % professional-quality tables
\usepackage{amsfonts}       % blackboard math symbols
\usepackage{nicefrac}       % compact symbols for 1/2, etc.
\usepackage{microtype}      % microtypography
\usepackage{xcolor}         % colors
\usepackage{enumitem}
\usepackage{xcolor}
\usepackage{enumitem}
\usepackage{xspace}
\usepackage{booktabs}
\usepackage{colortbl}
\usepackage{multirow}
\usepackage{makecell}
\usepackage{lipsum} % Just for generating dummy text, can be removed
\usepackage{gensymb} % for \degree
\usepackage{graphicx}
\usepackage{amssymb}
\usepackage{wrapfig}
\usepackage{bm}
\definecolor{MyDarkRed}{rgb}{0.66, 0.16, 0.16}
\definecolor{MyDarkBlue}{rgb}{0.16, 0.16, 0.66}

\usepackage{booktabs}   %% For formal tables:
\usepackage{subcaption} %% For complex figures with subfigures/subcaptions
\usepackage{algorithm}
\usepackage{xcolor}
\usepackage[noend]{algpseudocode}
\usepackage{algorithmicx}
\usepackage{appendix}
\usepackage{multirow}
\usepackage{diagbox}

\makeatletter
\DeclareRobustCommand\onedot{\futurelet\@let@token\@onedot}
\def\@onedot{\ifx\@let@token.\else.\null\fi\xspace}

\makeatother

\usepackage[labelsep=period]{caption}
\renewcommand{\paragraph}[1]{\vspace{0.05cm}\noindent \textbf{#1 \hspace{0.2em}}}

\PassOptionsToPackage{numbers, compress}{natbib}

%% file: sec/0_abstract.tex
\begin{abstract}
Significant progress has been made in spatial intelligence, spanning both spatial reconstruction and world exploration. However, the scalability and real-world fidelity of current models remain severely constrained by the scarcity of large-scale, high-quality training data. While several datasets provide camera pose information, they are typically limited in scale, diversity, and annotation richness, particularly for real-world dynamic scenes with ground-truth camera motion.
To this end, we collect \textbf{\DATAname}, a dataset consisting of a large corpus of in-the-wild videos with diverse scenes, camera movements and dense 3D annotations such as per-frame camera poses, depth, and motion instructions.
% providing the foundation for constructing a dataset with unique scale and diversity. 
% Specifically, we introduce \textbf{\DATAname}, a large-scale dynamic spatial dataset explicitly designed to provide expressive annotations for this purpose. 
Specifically, we collect more than 21,000 hours of raw video, and process them into 2.7 million clips through a hierarchical filtering pipeline, totaling 7,089 hours of dynamic content. A subsequent annotation pipeline enriches these clips with detailed spatial and semantic information, including camera poses, depth maps, dynamic masks, structured captions, and serialized motion instructions.
Analysis of SpatialVID's data statistics reveals a richness and diversity that directly foster improved model generalization and performance, establishing it as a key asset for the video and 3D vision research community.
Through extensive validation experiments, we demonstrate SpatialVID’s effectiveness across tasks such as controllable video generation, world simulation and geometric reconstruction, providing a strong foundation for spatial intelligence research.
\end{abstract}

%% file: sec/1_introduction.tex
% \begin{figure*}[htbp]
%     \centering
%     \begin{overpic}[width=.95\textwidth,
%     clip
%     ]{figure/vis/sample-main.pdf}
%     % \vspace{-8mm}
%     \end{overpic}
%   \caption{\textbf{Sample videos from \DATAname}.
%     Each example includes synchronized geometry, captions, and spatial annotations. The dataset encompasses diverse environments and camera motions, highlighting its broad coverage and multimodal consistency.
%     }
%   \label{fig:vid_sample}
% \end{figure*}

\section{Introduction}
\label{sec:intro}

\begin{table*} %[!h]
  % \vspace{-1mm}
  \caption{\textbf{Comparisons with previous datasets with spatial information.}
  \DATAname~is a \textit{million-level, dynamic and open-scenario} high-quality video dataset with rich annotated geometric and semantic information. 
  Syn. denotes synthetic data; Sta. and Dyn. indicate static and dynamic scenes.
  In \textit{Geometry Info.} column, C. denotes camera, D. denotes depth or point cloud.
  }
  \centering
  \label{tab:datasets_comp}
  \centering
  \resizebox{1.\linewidth}{!}{
  \begin{tabular}{lccccccc}
    \toprule
    \textbf{Dataset} & \textbf{Domain} & \textbf{Real/Syn.} & \textbf{Dyn./Sta.} & \textbf{Geometry Info.} & \textbf{Caption} & \textbf{\# Video Clips} & \# \textbf{Frames}
       % Annotations
    % \# Frames w/ 3D Anno.
    \\
    \midrule
    % 第一类：Syn.（先静态后动态）
    BlendedMVS~\citep{yao2020blendedmvs}
    & Open & Syn. & Sta. & C. D. & Label & 113 (Scenes) & 17K \\
    Multi-Cam Video~\citep{bai2025recammaster} & Open & Syn. & Dyn. & C. & Label & 136K & $\sim$11.02M \\
    PointOdyssey~\citep{zheng2023point} & Walk & Syn. & Dyn. & C. D. & N/A & 159 & 200K \\
    
    % 第二类：Real&Syn.（仅动态）
    Camerabench~\citep{lin2025towards} & Open & Real\&Syn. & Dyn. & N/A & Label, Short & 3,381 & - \\
    
    % 第三类：Real（先静态后动态）
    ScanNet~\citep{dai2017scannet} & Indoor & Real & Sta. & C. D. & N/A & 1500 (Scenes) & 2.50M \\
    MVImgNet~\citep{yu2023mvimgnet} & Object-Centric & Real & Sta. & C. & Label & 219,188 & 6.50M \\
    CO3Dv2~\citep{reizenstein2021common} & Object-Centric & Real & Sta. & C. & Label & 19K & 1.50M \\
    DL3DV~\citep{ling2024dl3dv} & Open & Real & Sta. & N/A & Label & 10,510 & 512M \\
    WebVi3D~\citep{ma2025you} & Open & Real & Sta. & N/A & - & 15.99M & 320M \\
    
    Dynpose100k~\citep{rockwell2025dynamic} & Open & Real & Dyn. & C. & Short & 100,131 & 6.81M \\
    CamVid-30K~\citep{zhao2024genxd} & Open & Real & Dyn. & C. & N/A & 30K & - \\
    Stereo4d~\citep{jin2024stereo4d} & Fisheye & Real & Dyn. & C. D. & Short & 110K & 10M \\
    RealEstate10K~\citep{zhou2018stereo} & Indoor & Real & Sta. & C. & N/A & $\sim$80K & $\sim$10M \\
    Waymo~\citep{sun2020scalability} & Drive & Real & Dyn. & C. & N/A & 1150 & 230K \\
    Map-free~\citep{arnold2022mapfree} & Object-Centric & Real & Sta. & C. & N/A & 655 (Scenes) & $\sim$560K \\
    Princeton365~\citep{princeton365} & open & Real & Dyn. & C. D. & N/A & 365 & $\sim$1.19M \\
    % Sekai~\citep{li2025sekai} & Open & Real & 
    \midrule
    \textbf{\DATAname} & Open & Real & Dyn. & C. D. & Structured Caption & 2.71M & 127.60M \\
    \textbf{\DATAname-HQ} & Open & Real & Dyn. & C. D. & Structured Caption & 0.37M & 20.63M \\
    
    \bottomrule
  \end{tabular}
  }
  \vspace{-3mm}
  
\end{table*}

% 1. spatial intelligence / world model development

% 2. video generation/understanding and 3d/4d generation/reconstruction are parts of spatial intelligence

% 3. deep dive into video generation development: (t2v i2v contrl-t2v)

% 4. 3d reconstruction developments

%  ----------------- 总写 -------------------------------------
% before start para.
% Perceiving, reasoning about, and interacting with the 3D world are crucial to artificial general intelligence.Recent advances has developed intrinsic 3D models that generate consistent, navigable environments from simple prompts, supporting both reconstruction and future simulation~\citep{genie3}.These representations naturally align with the dual objectives of recovering the current physical world and imagining plausible future worlds.
% tightened wjh 1105
Perceiving, reasoning about, and interacting with the 3D world are fundamental components of artificial general intelligence. Recent advances in intrinsic 3D modeling have enabled the generation of spatially consistent and navigable environments directly from text or image prompts, bridging reconstruction and world simulation~\citep{genie3}. These models embody the dual objectives of understanding the current physical world and imagining plausible future ones.
The field of 3D scene understanding has evolved from optimization-based methods to scalable, data-driven representations of geometry and motion. Classical approaches such as Structure-from-Motion (SfM)~\citep{Schonberger_2016_CVPR} and neural Multi-View Stereo (MVS)~\citep{yao2018mvsnet} laid the foundation for geometric reasoning. Subsequently, large-scale neural models have recently emerged: the LRM series~\citep{hong2023lrm,zhang2024gs,wei2024meshlrm} learns to reconstruct high-quality 3D objects from single images or text prompts; DUSt3R series~\citep{wang2024dust3r,cabon2025must3r} achieve robust multi-view matching; VGGT~\citep{wang2025vggt} directly predicts key 3D attributes such as camera poses and point clouds in a feed-forward manner. Collectively, these models exemplify a shift toward scalable, data-driven reconstruction and synthesis with reduced reliance on explicit geometry. Despite these advancements, building large-scale 3D datasets remains challenging due to costly data acquisition and the reliance on accurate annotation pipelines~\citep{yao2020blendedmvs, deitke2023objaverse, roberts2021hypersim}. In contrast, videos are abundant and naturally encode spatial, temporal, and semantic cues, offering a scalable foundation for 3D learning at scale. Leveraging such data presents a promising path toward high-fidelity reconstruction and dynamic world simulation.

Beyond reconstruction, video generation has emerged as a key capability for world modeling, serving as a simulator to represent and predict physical dynamics. Recent models including UNet-based diffusion methods like Stable Video Diffusion (SVD)~\citep{blattmann2023stable}, DiT-based architectures such as Sora~\citep{openai2024sora}, HunyuanVideo~\citep{kong2024hunyuanvideo}, and CogVideoX~\citep{yang2024cogvideox}, as well as autoregressive approaches~\citep{bardes2024revisiting} enable high-fidelity video synthesis. To achieve controllable and physically grounded generation, recent efforts extend to object motion~\citep{yin2023dragnuwa}, camera trajectory control~\citep{he2024cameractrl,wang2024motionctrl}, and 3D signal integration~\citep{yu2024viewcrafter}. These capabilities represent crucial steps toward physically grounded video simulation and align with recent world models such as Cosmos~\citep{agarwal2025cosmos}, HunyuanWorld~\citep{team2025hunyuanworld}, and Genie3~\citep{genie3}. Despite this progress, current datasets still lack detailed fine-grained semantic and spatial metadata, limiting their capacity to support physically grounded video synthesis.

Current data falls into two disjoint categories. Large-scale video datasets provide rich semantics but lack explicit 3D information~\citep{chen2024panda, nan2024openvid}, offering no geometric ground truth and forcing models to learn spatial relations implicitly from pixels. Conversely, spatial datasets such as CO3D~\citep{reizenstein2021common}, RealEstate10K~\citep{zhou2018stereo}, and Tartanair~\citep{wang2020tartanair} provide accurate geometry and camera parameters but remain small, object-centric, or synthetic. This division hampers progress toward spatiotemporally coherent world simulators and underscores the need for a dataset that bridges \textit{scene reconstruction} and \textit{world simulation}, closing the gap between semantic diversity without geometry and geometric precision without semantics.
% ours
% To bridge the gap between dynamic videos and spatial understanding, we introduce \textbf{SpatialVID} (\figref{fig:teaser}), a large-scale multimodal dataset that connects raw pixels with the physical world. Our curation pipeline begins with over 21,000 hours of raw internet video, manually screened to ensure diversity and motion richness. \wjh{\sout{A hierarchical filtering process (\figref{fig:data_flow}) distills the raw videos into a 7,089-hour core dataset of high-quality 720P clips with high-quality camera motion.} We then apply a hierarchical filtering process to distill the raw videos into a 7,089-hour core dataset of high-quality 720P clips with high-quality camera motion.} From this core, we construct \textbf{SpatialVID-HQ}, a 1,146-hour balanced subset optimized for robust model training and evaluation.
% To bridge the gap between dynamic videos and spatial understanding, we introduce \textbf{SpatialVID} (\figref{fig:teaser}), a large-scale multimodal dataset that connects raw pixels with the physical world. Our curation pipeline begins with over 21,000 hours of raw internet video, manually screened to ensure diversity and motion richness. We then apply a hierarchical filtering process to distill the raw videos into a 7,089-hour core dataset of high-quality 720P clips with high-quality camera motion. From this core, we construct \textbf{SpatialVID-HQ}, a 1,146-hour balanced subset optimized for robust model training and evaluation.
% wjh revise, 1105
To bridge the gap between dynamic videos and spatial understanding, we introduce \textbf{SpatialVID} (\figref{fig:teaser}), a large-scale multimodal dataset that connects raw pixels to the physical world. Starting from over 21,000 hours of internet video manually screened for diversity and motion richness, we apply a hierarchical filtering pipeline to obtain 7,089 hours of high-quality 720P clips with reliable camera motion. From this corpus, we derive \textbf{SpatialVID-HQ}, a 1,111-hour balanced subset optimized for robust training and evaluation. Through extensive experiments, we demonstrate that SpatialVID advances tasks such as controllable video generation, world simulation and geometric reconstruction, providing a strong foundation for spatial intelligence research.

To our knowledge, SpatialVID is the largest dataset of dynamic videos with explicit geometric annotations and makes three primary contributions:

% \begin{itemize}
%     \item \textbf{Manually Screened Videos with Camera Motions}: SpatialVID is built from a massive internet video collection of more than 21,000 hours, which are manually selected for rich scene motion. This motion-first curation, followed by our processing pipeline, yields a diverse set of high-quality clips well-suited for training spatially aware models.

%     \item \textbf{Comprehensive Geometric Annotations}: For each clip in the SpatialVID dataaset, we provide camera poses and depth maps generated via an adjusted pipeline~\citep{li2024_megasam}. These annotations supply explicit 3D grounding and motion dynamics, filling a key gap in existing video datasets.

%     \item \textbf{Spatially-Aware Captions and Motion Instructions}: We generate structured captions that integrate scene descriptions, camera motion details, and hierarchical semantic attributes such as weather, lighting, and time of day, providing rich text-video alignment. In addition, motion instructions are derived from camera trajectories, offering precise supervision for navigation-related model training.

% \end{itemize}

\begin{itemize}
    \item \textbf{Manually Screened Videos with Camera Motions}: SpatialVID is built from 21,000+ hours of internet videos manually screened for rich scene motion. This motion-first curation and processing yield diverse, high-quality clips for training spatially aware models.

    \item \textbf{Comprehensive Geometric Annotations}: Each clip includes camera poses and depth maps generated via an adjusted pipeline~\citep{li2024_megasam}, providing explicit 3D grounding and motion dynamics absent in prior video datasets.

    \item \textbf{Spatially-Aware Captions and Motion Instructions}: Structured captions describe scene content, camera motion, and semantic attributes (e.g., weather, lighting, time), while motion instructions derived from trajectories offer precise supervision for navigation and control tasks.

    % \wjh{\sout{add experiment or not?}\item \textbf{Experiments about Validation}}
\end{itemize}

%% file: sec/2_relatedwork.tex
\section{Related Work}
\label{sec:relatedwork}

\paragraph{Scene Reconstruction}
Traditionally, scene reconstruction has followed two main trajectories: Simultaneous Localization and Mapping (SLAM) and Structure-from-Motion (SfM). Classical systems such as ORB-SLAM~\citep{mur2015orb} and COLMAP~\citep{Schonberger_2016_CVPR} achieve accurate geometry estimation and real-time tracking but depend heavily on handcrafted features, limiting robustness under challenging conditions. To overcome these constraints, learnable Multi-View Stereo (MVS) methods have emerged, with MVSNet~\citep{yao2018mvsnet} introducing deep cost-volume reasoning and Transformer-based models such as DUSt3R~\citep{wang2024dust3r}, MASt3R~\citep{leroy2024grounding}, and their successors~\citep{cabon2025must3r,lu2025align3r,yang2025fast3r} achieving robust feed-forward multi-view matching. Recent frameworks like VGGT~\citep{wang2025vggt} further integrate these advances into end-to-end 3D reconstruction pipelines. 
Extending this trend to dynamic environments, recent DUSt3R variants~\citep{zhang2024monst3r,chen2025easi3r,cut3r} incorporate motion awareness extend the capabilities of DUSt3R to incorporate dynamic elements, while dense approaches including CasualSAM~\citep{zhang2022structure} and MegaSaM~\citep{li2024_megasam} leverage optical flow or SLAM backbones~\citep{teed2021droid} for robust tracking. Given its demonstrated robustness in unconstrained, in-the-wild videos, we utilize an enhanced version of MegaSaM in our work to generate the initial geometric annotations for our dataset.

\paragraph{World Simulator}
% video generation
The concept of a world simulator lies at the core of spatial intelligence, encompassing the ability to perceive, simulate, and interact with dynamic environments. Recent progress in video generation has made this increasingly feasible. Early progress was driven by UNet-based video diffusion models such as Stable Video Diffusion (SVD)~\citep{blattmann2023stable}, followed by DiT-based architectures~\citep{openai2024sora,kong2024hunyuanvideo,yang2024cogvideox}, which significantly advance fidelity, scalability, and temporal consistency. In parallel, autoregressive approaches~\citep{bardes2024revisiting,yu2025context} remain competitive for generating high-quality, long-form videos. Achieving effective world simulation further requires controllability over 3D structure and motion. 
Models such as DragNUWA~\citep{yin2023dragnuwa}, CameraCtrl~\citep{he2024cameractrl}, and MotionCtrl~\citep{wang2024motionctrl} enable explicit manipulation of objects and camera trajectories, while GameFactory~\citep{yu2025gamefactory} introduces action-level control for interactive synthesis. Incorporating 3D data further enhances spatial coherence, as demonstrated by ViewCrafter~\citep{yu2024viewcrafter}, which integrates point clouds into video generation. These controlled and geometry-aware paradigms contribute to the development of world models such as Cosmos Predictor~\citep{agarwal2025cosmos}, HunyuanWorld~\citep{team2025hunyuanworld}, and Genie3~\citep{genie3}, aiming to simulate spatiotemporal dynamics and support interactive exploration within complex virtual environments.

\begin{figure*}[t]
    \centering
    \begin{overpic}[width=1.\textwidth,
    % \begin{overpic}[width=1.\columnwidth,
    % trim=20 100 20 50, 
    clip
    ]{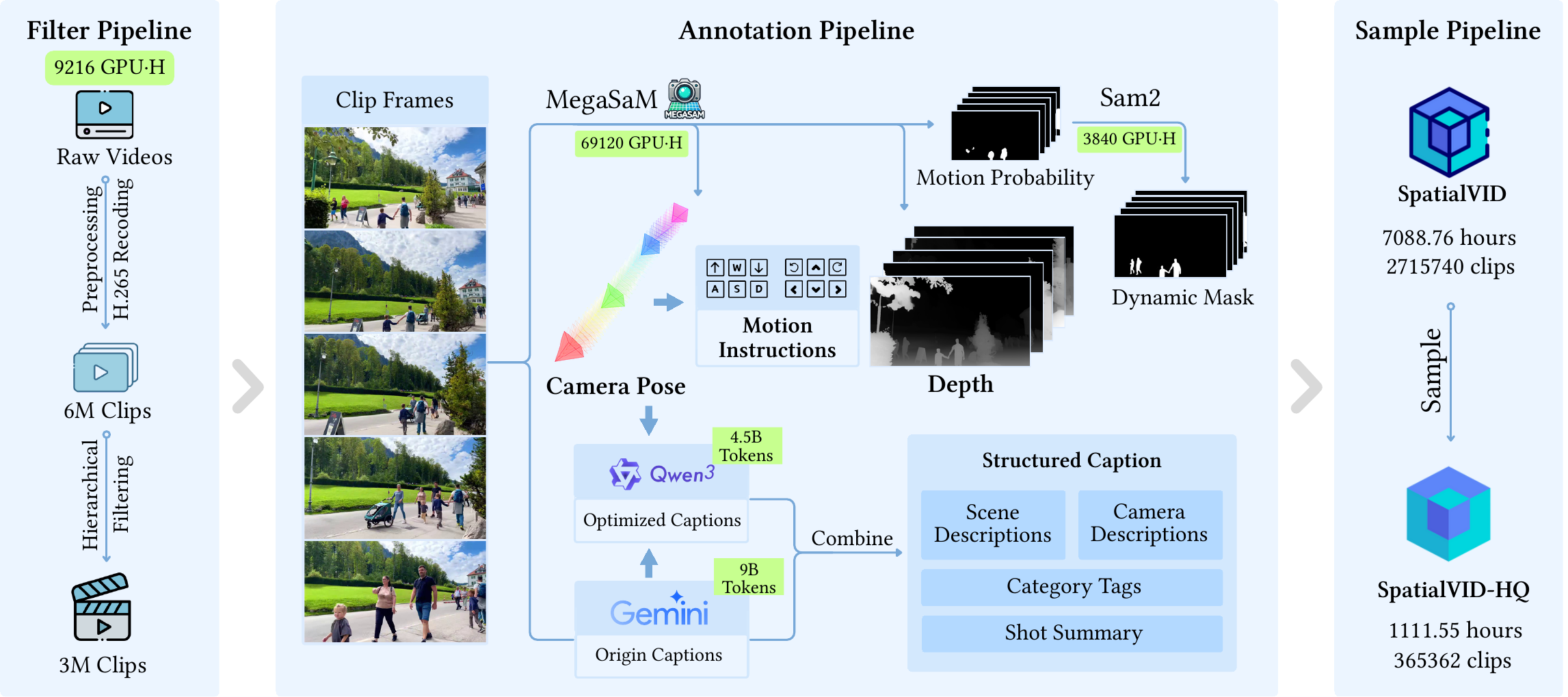}
    \end{overpic}
% \caption{\textbf{Overview of the curation pipeline}. The pipeline comprises three stages: filtering, annotation, and sampling. It begins with raw web videos manually collected for camera motion, yielding high-quality dynamic clips with geometric and semantic annotations. In the \textbf{filtering stage}, raw videos are preprocessed and filtered hierarchically. 
% The \textbf{annotation stage} performs geometric and semantic labeling while deriving motion instructions from camera poses. Finally, in the \textbf{sampling stage}, clips are balanced by motion characteristics and categories to generate a high-quality (HQ) subset with evenly distributed classes, ensuring comprehensive coverage for downstream tasks.
% }
\caption{\textbf{Overview of the curation pipeline.} The pipeline comprises three stages: filtering, annotation, and sampling. We start from manually collected web videos with notable camera motion. In the \textbf{filtering stage}, raw videos are hierarchically preprocessed and filtered. The \textbf{annotation stage} adds geometric and semantic labels and derives motion instructions from camera poses. The \textbf{sampling stage} then balances clips by motion and category to form a high-quality subset (SpatialVID-HQ) with well-distributed classes for downstream tasks.}
  \label{fig:pipe}
  \vspace{-2mm}
\end{figure*}

\paragraph{Datasets with Spatial Annotations}
High-quality spatial datasets are crucial for advancing world reconstruction and exploration, ideally combining 3D geometry, dynamics, and semantic richness. Existing efforts fall into three main categories. Synthetic datasets such as Multi-Cam Video~\citep{bai2025recammaster}, virtualKITTI~\citep{cabon2020virtual}, and BlendedMVS~\citep{yao2020blendedmvs} offer precise geometry information but demand heavy engineering, limiting scalability. Real-world datasets like CO3DV2~\citep{reizenstein2021common} and RealEstate10K~\citep{zhou2018stereo} rely on SfM or SLAM-based annotation for efficiency, often yield sparse trajectories and struggle under dynamic motion. Video-mined datasets including CamVid-30K~\citep{zhao2024genxd} and DynPose100K~\citep{rockwell2025dynamic} provide large-scale semantic or motion cues but often lack fine-grained geometry or diversity. Recent datasets including CameraBench~\citep{lin2025towards}, VLM4D~\citep{zhou2025vlm4d}, and the concurrent Sekai~\citep{li2025sekai} improve geometric–semantic integration but remain limited in semantic richness, motion coverage, or scale. To address these limitations, we introduce \DATAname, a large-scale real-world dataset of diverse dynamic scenes with detailed geometric and semantic annotations. As shown in \tabref{tab:datasets_comp}, \DATAname~outperforms prior datasets in scale and annotation quality, providing a rich foundation for spatially grounded representation learning and world simulation research.

%% file: sec/3_curation.tex
\section{\DATAname~Curation}

% \wjh{overview of paragraph}
% As illustrated in \figref{fig:pipe}, the SpatialVID data processing pipeline is organized into three core stages: filtering, annotation, and sampling, each of which is elaborated in the following subsections. 
% \secref{sec:curation:subsec:preprocess} describes the video collection procedure, the unification of encoding formats, and the initial segmentation of clips using \texttt{scenedetect}. \secref{sec:curation:subsec:filter} introduces the multi-dimensional filtering strategy based on aesthetic quality, motion intensity, OCR-based text interference, and luminance, with the goal of retaining clips that contain diverse and meaningful motion. 
% \secref{sec:curation:subsec:slam} explains the annotation of geometric camera motion, including the generation of camera poses and depth maps that serve as spatial priors for each clip. 
% \secref{sec:curation:subsec:instruction} details the creation of motion instructions, and 
% \secref{sec:curation:subsec:caption} describes the annotation of structured captions, which integrate scene descriptions, camera motion details, and other semantic labels such as weather, lighting, and time of day. 
% Representative examples of motion instruction annotations are shown in \figref{fig:instruction_vis}, while additional results illustrating geometric and semantic annotations are also presented in \figref{fig:vid_sample}.
As illustrated in \figref{fig:pipe}, the SpatialVID curation pipeline comprises three main stages: filtering, annotation, and sampling.
\secref{sec:curation:subsec:preprocess} covers video collection, format unification, and clip segmentation;
\secref{sec:curation:subsec:filter} introduces multi-dimensional filtering based on four key metrics to retain diverse, motion-rich clips;
\secref{sec:curation:subsec:slam} describes geometric annotation via camera poses and depth maps;
\secref{sec:curation:subsec:instruction} details motion instruction generation; and
\secref{sec:curation:subsec:caption} outlines structured captions integrating scene, motion, and semantic attributes.
% Representative motion annotations appear in \figref{fig:instruction_vis}, with additional examples in \wjhref{Appendix specific section \figref{fig:vid_sample}}.

% 在右侧展示图片，figure\statistics\captions\raw_video_dist.png，并且正文在图片左侧
% \begin{wrapfigure}{htbp}{0.38\textwidth} % htbp
%     \vspace{-5mm}
%     \centering
%     \includegraphics[width=0.38\textwidth]{figure/statistics/tag_dist/raw_video_dist.pdf}
%     % \caption{\textbf{Distribution of raw video duration.} The distribution of raw video duration collected from YouTube. Most videos are between 1 and 10 minutes, with a few exceeding 30 minutes.}
%     \caption{\textbf{Distribution of raw video duration.}
%       % In the pie chart of video shooting camera carriers, a rich variety of shooting scenarios and carrier types are covered: videos shot indoors are uniformly named house tour; the camera carriers in outdoor scenarios are more diverse, including walking, train, drone, and other types. It can be seen that the dataset has a wide coverage of shooting carrier categories.
%       The left panel shows the quantity distribution, while the right panel presents the duration distribution. The charts show diverse scenarios and carriers: indoor (house tour) and outdoor (walking, train, drone, etc.), reflecting broad coverage of shooting contexts.
%     }
%     \label{fig:raw_video_dist}
%     \vspace{-5mm}
% \end{wrapfigure}

% 尽管一些工作是从现有视频数据集中进行筛选的

\subsection{Data Collection and Preprocessing}
\label{sec:curation:subsec:preprocess}

\noindent \textbf{Source Data.} While large-scale video datasets such as Panda70M~\citep{chen2024panda} and MiraData~\citep{ju2024miradatalargescalevideodataset} offer valuable benchmarks, they remain inadequate for our needs due to limited viewpoints and motion diversity. Processing the Panda70M validation split through our pipeline retained only about 10\% of clips that met our quality standards, yielding insufficient data for large-scale reconstruction (\secref{sec:analysis:subsec:compare}).

To obtain richer and more varied data, we turned to YouTube, leveraging its vast and heterogeneous collection of high-resolution videos. We queried motion-related keywords (\textit{walk}, \textit{tour}, \textit{drone}, etc.) to capture videos with smooth and diverse camera trajectories.
Candidate videos were then manually screened for compatibility with the MegaSaM reconstruction pipeline. We excluded broken videos and videos with titles containing inappropriate terms (e.g., "Panoramic camera"). Footage with heavy occlusion or intrusive overlays (e.g., logos, subtitles) was also discarded.

This process yielded a curated corpus with stable motion, rich parallax, and detailed textures, ideal for 3D reconstruction. In total, we collected 33,443 YouTube videos (21,789 hours), covering a broad range of motion types, camera trajectories, and scene categories. Refer to Supplementary Materials for additional details about video statistics.

\noindent \textbf{Data Preprocessing.}
We segment the collected long-form videos into 3–15 s clips using a modified PySceneDetect~\citep{PySceneDetect}. To better handle aesthetic transitions such as fades, we adjust sensitivity thresholds and adopt an interval-based multi-frame comparison strategy, significantly improving segmentation accuracy and efficiency. Given the variability in encoding formats and resolutions, all clips are then standardized to H.265-encoded MP4 at $1280\times720$ to ensure consistency and compatibility across the pipeline. After preprocessing, this yields over 7 million distinct video clips.

\subsection{Video Quality Filtering}
\label{sec:curation:subsec:filter}

We filter all candidate videos using four key metrics: aesthetic quality, motion intensity, text interference, and luminance. This stage ensures that only clips with rich motion and clear visuals are retained, improving the reliability of downstream camera pose estimation. Given the large scale of our collection, such filtering is crucial for maintaining dataset quality and suitability for training and evaluation.

\begin{figure} % [h!]
    % \vspace{-18mm}
    \centering
    \begin{overpic}[width=\columnwidth,
    % height=6cm,
    % trim=40 140 20 90,
    clip
    ]{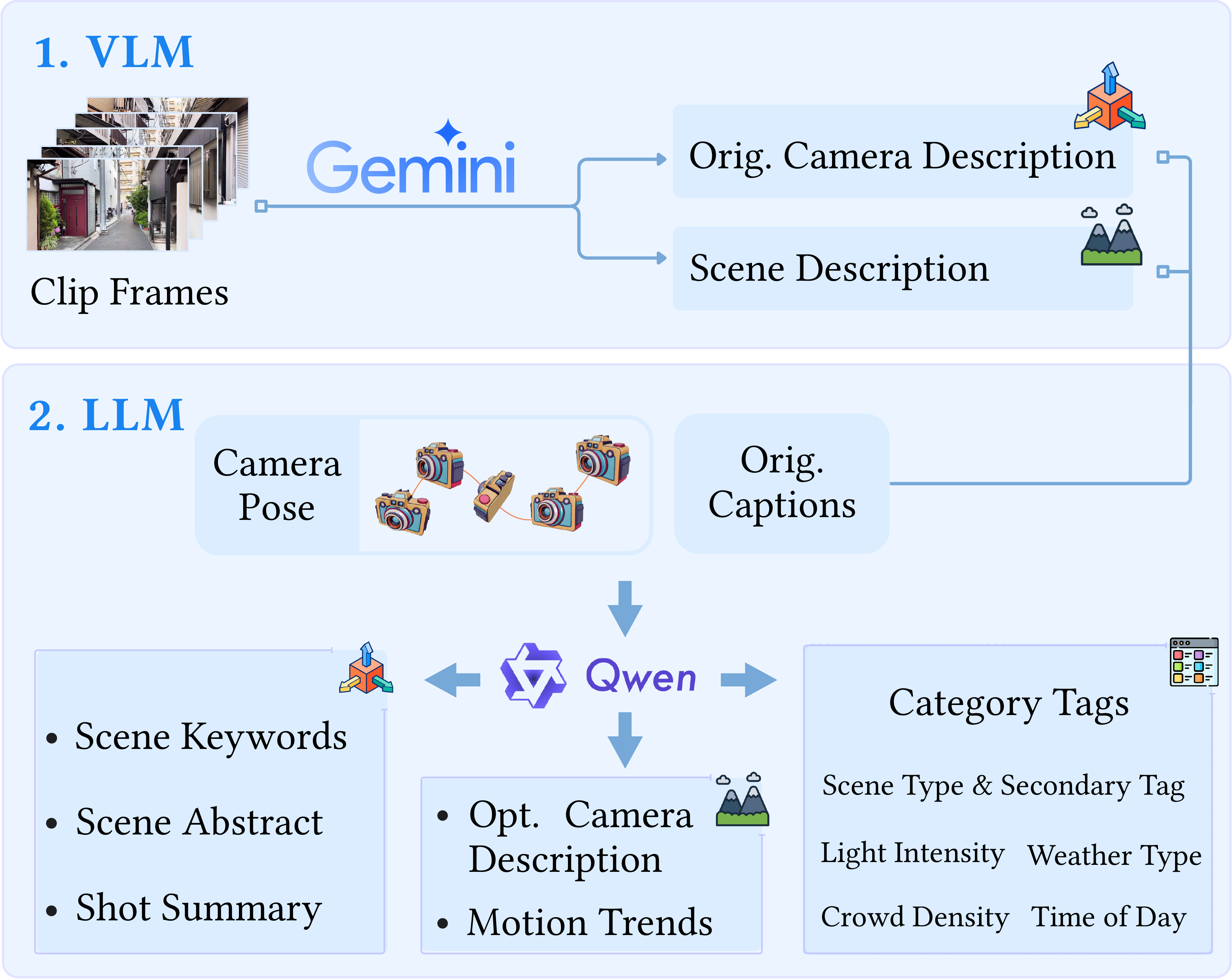}
    % \vspace{-8mm}
    \end{overpic}
    \caption{\textbf{Structured caption generation.} A VLM produces initial motion and scene descriptions, which the LLM refines using camera poses to yield structured captions.}
        \label{fig:caption_pipe}
    \vspace{-3mm}
\end{figure}

Specifically, we adopt a CLIP+MLP aesthetic predictor~\citep{schuhmann2022laion} to remove visually unappealing clips. Luminance filtering discards overexposed or underexposed samples to ensure consistent brightness. PaddleOCR~\citep{cui2025paddleocr} is used to detect and eliminate clips with excessive text based on the text-area ratio. Finally, motion filtering leverages lightweight VMAF metric~\citep{VMAF2016} to retain videos with sufficient motion diversity for stable reconstruction. Representative filtering examples are provided in Supplementary Materials.

% \textbf{Aesthetic Filtering}. To quantitatively assess visual appeal, we use a CLIP+MLP aesthetic score predictor~\citep{aesthetic_predictor}. The model assigns a score from 0 to 10, with higher values indicating better quality. For each video clip, the score is averaged across the first, middle, and last frames. Clips with an average score below 4.0 are considered insufficiently appealing and discarded \wjh{\sout{(\figref{fig:aes})}}.

% \textbf{Luminance Filtering}. Luminance is calculated for the first, middle, and last frames using the standard formula \(L = 0.2126 R + 0.7152 G + 0.0722 B\), where \(R\), \(G\), and \(B\) are the respective channel values. Clips with average luminance outside the range [20, 140], either too dark or too bright, are excluded, ensuring that only videos with proper exposure are retained\wjh{ \sout{(\figref{fig:lum})}}.

% \textbf{Optical Character Recognition (OCR)}. For text detection, we use the latest release of PaddleOCR~\citep{paddleocr2020}, which offers high accuracy and robust multilingual support. We processe the first, middle, and last frames of each clip to detect text regions, computing the ratio of text area to frame size. Clips where the text area exceeded 30\% are removed, as these are considered informational rather than visual \wjh{\sout{(\figref{fig:ocr})}}.

% \textbf{Motion Filtering}. We use lightweight VMAF~\citep{VMAF2016}, integrated FFmpeg with the valid motion score ranging from 2.0 to 14.0 \wjh{\sout{(\figref{fig:motion})}}.

\subsection{Geometry Information Annotation}
\label{sec:curation:subsec:slam}

We employ MegaSaM~\citep{li2024_megasam} as our primary camera estimator, chosen for its strong balance between accuracy and efficiency compared to existing methods~\citep{teed2021droid,schoenberger2016sfm,yang2025fast3r,zhang2024monst3r,wang2025vggt}. Comparisons of several methods are presented in Supplementary Materials. Our annotation pipeline is poised to improve by integrating future estimators with enhanced performance (e.g., ViPE~\citep{huang2025vipe}).

While effective, MegaSaM faces challenges under extreme conditions, such as dominant moving objects, collinear motion, or reliance on external monocular depth. To address these issues, we replace its original depth module with UniDepth v2~\citep{piccinelli2025unidepthv2} and Depth Anything v2~\citep{depth_anything_v2}, significantly improving depth accuracy and robustness.
Furthermore, we first obtain candidate regions via adaptive thresholding and contour detection, then sample anchor points from them and use these as SAM2~\citep{ravi2024sam2} prompts to extract dynamic masks.
From these refined masks, we compute \textit{dynamic ratio} that quantifies the proportion of dynamic regions per frame.

To ensure physically plausible trajectories, we employ an acceleration-based detector to identify abrupt, non-physical motion fluctuations. Additionally, we introduce three metrics for quantitative camera motion analysis: (1) \textit{MoveDist}, measuring total trajectory length; (2) \textit{RotAngle}, capturing cumulative camera rotation; and (3) \textit{TrajTurns}, estimating the number of significant directional changes.

\begin{figure}% [h!]
    % \vspace{-18mm}
    \centering
    % \begin{overpic}[width=1.\textwidth,
    \begin{overpic}[width=\columnwidth,
    % height=6cm,
    % trim=40 140 20 90,
    clip
    ]{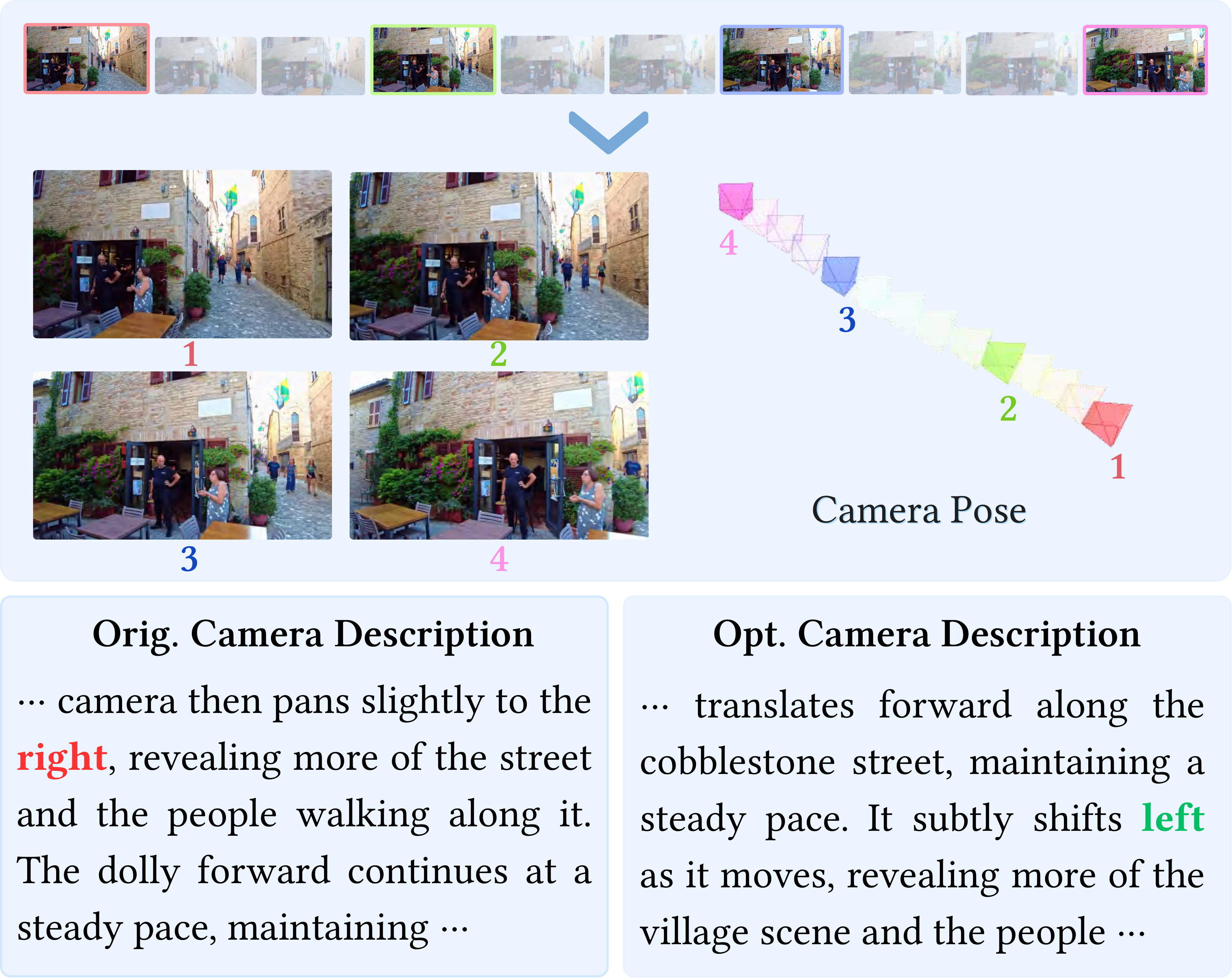}
    % \vspace{-8mm}
    \end{overpic}
    \caption{\textbf{Effect of spatial enhancement.} After applying spatial enhancement, the LLM corrected the incorrect direction (right) output by the VLM to left.}

        \label{fig:vlm_comp}
    \vspace{-3mm}
\end{figure}

\subsection{Motion Instruction Decomposition}
\label{sec:curation:subsec:instruction}
% before version in 2 para.
% \citep{li2025hunyuangamecrafthighdynamicinteractivegame}, our dataset explicitly incorporates them to facilitate controllable and semantically meaningful motion learning. To generate these motion instructions, we begin by processing the output sequences of camera poses estimated from each clip. The relative translations and rotations between consecutive frames serve as the basis for instruction derivation, as they directly encode camera motion dynamics. To ensure reliability, we apply temporal smoothing filters to the raw camera poses, effectively suppressing jitter and measurement noise that could otherwise result in spurious motion labels. 

% Next, we employ a thresholding mechanism to identify segments with perceptible motion: only when the magnitude of relative pose change between adjacent frames surpasses predefined thresholds do we generate an instruction. This step prevents the generation of trivial or redundant instructions in near-static scenarios. Finally, to maximize clarity and consistency, we map these motion signals to a controlled vocabulary of cinematographic terms \citep{deguzman2025types}, such as \textit{dolly in} (forward translation), \textit{pan left} (horizontal rotation to the left), and \textit{truck right} (lateral translation to the right), as exemplified in~\figref{fig:instruction_vis}. This decomposition not only standardizes the representation of camera motion but also makes it interpretable and readily usable for downstream model training, where high-level motion semantics are crucial.

\begin{figure*} % [t!]
    \centering
    \begin{overpic}[width=1.\textwidth,
    % trim=20 100 20 50, 
    clip
    ]{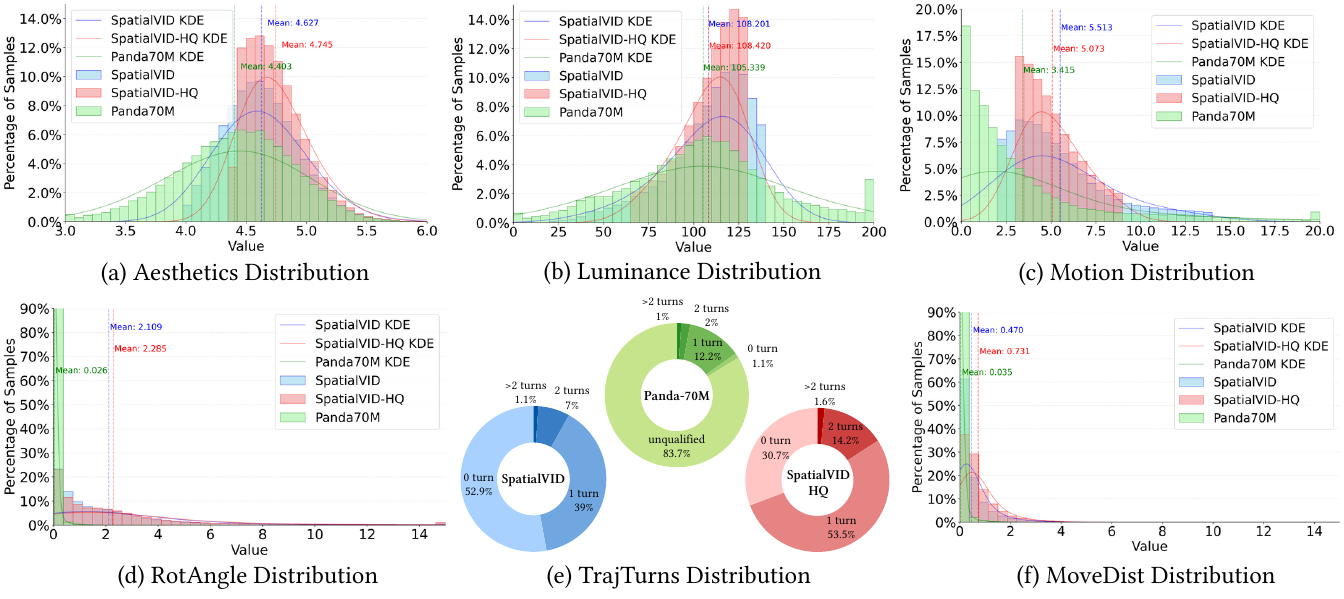}
    \end{overpic}
    %  \caption{
    %     \textbf{Dataset quality comparison.} A comparative analysis of our \DATAname, its high-quality subset (\DATAname-HQ), and the Panda70M-test set processed with our identical pipeline, visualized via histograms and Kernel Density Estimation (KDE) curves. KDE curves illustrate the continuous distribution patterns of each dataset, facilitating intuitive comparison of their value distribution shapes. The results demonstrate that \DATAname-HQ exhibits markedly superior quality across all metrics, validating the effectiveness of our manual collection, filtering and sampling methodology.
    % }
    \caption{\textbf{Dataset quality comparison.} Comparison of \DATAname, its balanced subset (\DATAname-HQ), and the Panda70M-test set processed with the same pipeline. Histograms and KDE curves reveal distribution patterns across quality metrics, showing that \DATAname-HQ achieves consistently superior quality, validating our manual collection, filtering, and sampling process.}
  \label{fig:comparison}
  \vspace{-2mm}
\end{figure*}

% wjh 1105
Given the importance of motion instructions for controllable and semantically grounded learning in models such as Hunyuan-GameCraft~\citep{li2025hunyuangamecrafthighdynamicinteractivegame}, our dataset explicitly incorporates them to enable interpretable motion understanding. We derive these instructions from estimated camera pose sequences, where relative translations and rotations between consecutive frames capture the camera’s motion dynamics. To enhance robustness, temporal smoothing filters are applied to suppress jitter and noise prior to instruction extraction. Perceptible motion segments are then identified through magnitude-based thresholding, generating instructions only when pose variations exceed predefined limits to avoid trivial movements. Finally, motion signals are mapped to a controlled vocabulary of cinematographic terms~\citep{lin2025towards}, such as \textit{dolly in} (forward translation), \textit{pan left} (horizontal rotation), and \textit{truck right} (lateral translation). Visualized annotation results are provided in Supplementary Materials. This standardized decomposition ensures clarity, consistency, and semantic usability for downstream training.

\subsection{Semantic Information Annotation}
\label{sec:curation:subsec:caption}

Multimodal models have made remarkable strides in bridging vision and language, employing diverse strategies for video caption generation. Early efforts such as VATEX~\citep{wang2019vatex} rely on manual annotation to ensure linguistic precision but lack scalability. More recent datasets~\citep{chen2024panda, tan2024vidgen, nan2024openvid} leverage multimodal large language models (MLLMs) for automatic caption generation, striking a balance between efficiency and semantic relevance. Others adopt structured or multi-stage pipelines~\citep{ju2024miradatalargescalevideodataset, wang2025koala} to improve vision–text alignment, while OpenHumanVid~\citep{li2025openhumanvid} enhances caption reliability through LLM-based voting and reformatting. Despite these advances, vision–language models (VLMs) such as Gemini~\citep{team2023gemini} remain limited in spatial reasoning, often overlooking geometric and camera-related details (\figref{fig:vlm_comp}). Recent works like CameraBench~\citep{lin2025towards}, VLM4D~\citep{zhou2025vlm4d}, and 3D LLM-Mem~\citep{hu20253dllm} highlight these shortcomings and explore early attempts to integrate depth and pose awareness. 

Building upon these insights, we propose a structured caption generation framework that unifies VLM and LLM capabilities while explicitly incorporating camera poses (\figref{fig:caption_pipe}), enabling captions that are both semantically expressive and spatially grounded. Our pipeline operates in two stages: (1) \textit{visual parsing}, where Gemini-2.0-Flash analyzes sampled frames to produce initial scene and motion descriptions; and (2) \textit{language refinement}, where Qwen3-30B-A3B~\citep{yang2025qwen3} refines these outputs using camera pose priors to correct motion directions and ensure spatial coherence. The refined captions integrate scene semantics, camera motion, and multi-level attributes (e.g., scene type, lighting, weather), forming a hierarchical textual representation that captures spatial, temporal, and cinematic structure. Further implementation details are provided in Supplementary Materials.

%% file: sec/4_analysis.tex
\section{Dataset Analysis}

\subsection{Data Sampling}
% Our primary objective is to curate clips that are both high quality and diverse. To achieve this, we adopt a two-step strategy. First, we raise the thresholds of key video quality metrics to improve the overall standard of the dataset. Second, we balance the distribution of semantic tags and camera trajectory statistics (e.g., Arc Nums) to preserve diversity across content types. This refined sampling procedure yields over one thousand hours of high-quality clips, forming our curated spatial video dataset, \DATAname-HQ.
% wjh revise 1105
We aim to curate clips that maximize both quality and diversity. To this end, we first tighten thresholds across core quality metrics to elevate visual standards, and then balance semantic tags and trajectory statistics to ensure diversity. The resulting collection, \DATAname-HQ, comprises over one thousand hours of high-quality spatial videos.

\subsection{Comparison with Panda-70M}
\label{sec:analysis:subsec:compare}
Panda-70M is a large-scale video dataset designed to support research on vision-language and video understanding. However, it exhibits several quality limitations: many clips are static, contain flickering artifacts, and include captions that lack motion descriptions. To demonstrate the advantages of our dataset, we perform a systematic comparison between \DATAname~and Panda-70M, as illustrated in \figref{fig:comparison}.

\begin{table*} %[h!]
\caption{
    Quantitative comparison of camera-controlled video generation performance across different training datasets on Sekai-Real~\citep{li2025sekai}, RealEstate10K\citep{zhou2018stereo}, and SpatialVID benchmarks.
}
\centering
\resizebox{1.\textwidth}{!}{
    \begin{tabular}{lccccccccccc}
    \toprule[1.2pt]
       \multirow{3}{*}{Benchmark} & \multirow{3}{*}{Training Data} 
    & \multicolumn{3}{c}{Camera Accuracy} & \multicolumn{2}{c}{Visual Quality} & \multicolumn{5}{c}{VBench} \\
    \cmidrule(lr){3-5} \cmidrule(lr){6-7} \cmidrule(lr){8-12} 
    & & TransErr$\downarrow$ &  RotErr$\downarrow$ & CamMC$\downarrow$ & CLIP-T$\uparrow$ & CLIP-F $\uparrow$  &  \makecell{Subject\\Consistency} & \makecell{Background\\Consistency} & \makecell{Motion\\Smoothness} & \makecell{Aesthetic\\Quality} & \makecell{Imaging\\Quality}  \\

    \midrule
    \multirow{3}{*}{RE10K} & RE10K & \underline{7.46} & \underline{1.15} & \underline{7.91} & 30.38 & \underline{99.53} & 96.45 & \underline{94.52} & \textbf{99.05} & \underline{55.99} & \underline{72.71} \\
    & Sekai-Real & 8.50 & 1.86 & 9.35 & \underline{30.51} & 99.20 & \underline{97.09} & 94.21 & 98.57 & 52.17 & 72.02 \\
    & \textit{SpatialVID-HQ}  & \textbf{7.42} & \textbf{0.99} & \textbf{7.72} & \textbf{30.54} & \textbf{99.59} & \textbf{98.04} & \textbf{95.41} & \underline{98.77} & \textbf{56.26} & \textbf{75.68} \\
    \midrule
     \multirow{3}{*}{Sekai}  & RE10K & 8.17 & 1.51 & 8.78 & \underline{34.97} & \textbf{99.35} & \underline{96.25} & \underline{93.24} & \textbf{99.12} & \textbf{54.91} & 71.27 \\
     & Sekai-Real & \underline{6.49} & \underline{1.47} & \underline{7.12} & 33.98 & 98.57 & 93.25 & 91.24 & 98.06 & 52.53 & \underline{72.32} \\
    & \textit{SpatialVID-HQ} & \textbf{6.04} & \textbf{1.43} & \textbf{6.70} & \textbf{35.19} & \underline{99.28} & \textbf{96.39} & \textbf{93.49} & \underline{98.88} & \underline{54.14} & \textbf{73.13} \\
    \midrule
    \multirow{3}{*}{SpatialVID} & RE10K  & \underline{5.16} & \underline{4.07} & \underline{8.59} & 30.22 & \textbf{99.02} & \underline{94.31} & \underline{92.70} & \textbf{98.68} & \underline{55.00} & \underline{66.23} \\
    & Sekai-Real & 5.63 & 4.70 & 9.39 & \underline{30.25} & 98.13 & 91.63 & 91.93 & 97.97 & 52.90 & 66.14 \\
    & \textit{SpatialVID-HQ} & \textbf{4.33} & \textbf{3.81} & \textbf{7.57} & \textbf{30.26} & \underline{98.69} & \textbf{94.89} & \textbf{93.23} & \underline{98.18} & \textbf{55.11} & \textbf{70.38} \\
    \bottomrule[1.2pt]
    \end{tabular}
}
\label{tab:cvd_compare}
  % \vspace{-3mm}
\end{table*}

% % vbench
% \begin{table*}
% \caption{
%     Quantitative comparison across different training datasets on VBench\citep{huang2024vbench} metrics.
% }
% \centering
% \resizebox{.95\textwidth}{!}{
% \begin{tabular}{lccccccc}
%     \toprule[1.2pt]
%     Method & \makecell{Subject\\Consistency} & \makecell{Background\\Consistency} & \makecell{Motion\\Smoothness} & \makecell{Aesthetic\\Quality} & \makecell{Imaging\\Quality} & \makecell{Temporal\\Flickering} & \makecell{Overall\\Consistency} \\
%     \midrule
%     Fine-tuned with Sekai & 94.33\% & 96.05\% & 99.05\% & 53.79\% & 64.84\% & 98.52\% & 18.94\% \\
%     Fine-tuned with Re10k & 96.94\% & 97.21\% & 99.28\% & 54.29\% & 64.74\% & 99.13\% & 19.36\% \\
%     Fine-tuned with SpatialVID-HQ & 91.31\% & 95.17\% & 98.08\% & 53.48\% & 60.94\% & 96.13\% & 20.29\% \\
%     \bottomrule[1.2pt]
% \end{tabular}
% }
% \label{tab:vbench}
%   \vspace{-1mm}
% \end{table*}

% qualitative
\begin{figure*}%[h!]
    \centering
    \begin{overpic}[width=1.\textwidth,
    % \begin{overpic}[width=1.\columnwidth,
    clip
    ]{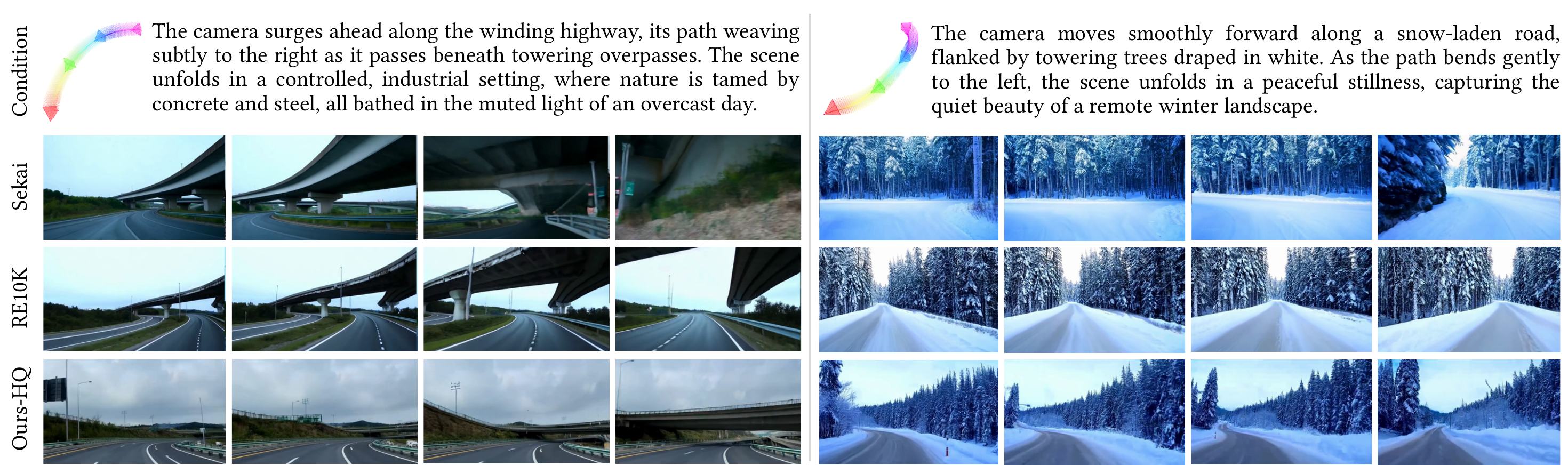}
    \end{overpic}
    \caption{\textbf{Different training datasets performance on SpatialVID.} Under identical training settings, models trained on our dataset produce videos with more consistent appearance and significantly improved camera controllability.}

    \label{fig:cvd_qualitative}
  \vspace{-2mm}
\end{figure*}
% \wjh{fig about camera control}

Across key video-quality metrics including Aesthetics (\figref{fig:comparison}a), Luminance (\figref{fig:comparison}b), and Motion (\figref{fig:comparison}c), both \DATAname~and its high-quality subset \DATAname-HQ show more compact distributions, indicating greater consistency and higher average quality. Regarding camera motion statistics, Panda-70M is dominated by static viewpoints, as seen in the distributions of camera rotation (\figref{fig:comparison}d) and translation distance (\figref{fig:comparison}f). In contrast, \DATAname~achieves a balanced and realistic distribution of camera movements. Finally, \figref{fig:comparison}e shows the distribution of trajectory turns (Arc Nums): over 80\% of Panda-70M videos cannot be reconstructed by MegaSaM due to insufficient motion, whereas \DATAname-HQ deliberately increases the proportion of clips with curved or turning trajectories, providing a richer and more diverse motion profile.

%% file: sec/5_task1.tex
\vspace{-1mm}

\section{SpatialVID Validation Tasks}
% 我们做了多种实验来验证数据集的有效性。
% 分别说明两个/三个实验的角度，与核心内容
% 说明下通过实验结果证明了数据集的有效性

% We select 2/3 tasks: Camera controlled video generation for world exploration, large reconstruction model for spatial reconstruction, and pose estimation models for data accuracy validation. % pose estimation这里可以参考下别的文章中怎么样写的
% In this section, we validate the effectiveness of the \DATAname~dataset through three core tasks: (1) camera-controlled video generation for world exploration, (2) large-scale spatial reconstruction, and (3) pose estimation for data accuracy assessment. These tasks jointly evaluate complementary aspects of the dataset and demonstrate its utility for advancing 3D environment understanding and generation.
% update
In this section, we primarily evaluate the effectiveness of the \DATAname~dataset on camera-controlled video generation, which serves as our main task to assess spatially consistent visual synthesis. In addition, we conduct large-scale spatial reconstruction and pose estimation experiments as complementary validations, providing further evidence of the dataset’s quality and geometric fidelity.

\subsection{Camera-Controlled Video Generation}

\noindent \textbf{Baseline.}
We build upon the camera-control injection mechanism introduced in ReCamMaster~\citep{bai2025recammaster} (excluding intrinsic parameters), adopting the Wan2.2 architecture~\citep{wan2025wan} as the base model and T5~\citep{raffel2020exploring} as the text encoder. For each frame, camera extrinsics are processed by a learnable encoder $E_c$ that projects them into the video token dimension. The encoded features are then fused with visual tokens and injected into each transformer block, enabling precise, frame-level camera control.
We train models on Sekai-Real~\citep{li2025sekai}, RealEstate10K, and our SpatialVID-HQ subset under identical training configurations. Each model is fine-tuned on 32 H20 GPUs for 2 days and detailed training setup is provided in Supplementary Material.

% 1113
\noindent \textbf{Metrics.}
Following prior works~\citep{he2024cameractrl,zheng2024cami2v}, we evaluate camera controllability using \textbf{TransErr}, \textbf{RotErr}, and \textbf{CamMC}, computed on poses estimated by MegaSaM~\citep{li2024_megasam}.
For a fair and comprehensive comparison, all methods are evaluated on 100 randomly sampled sequences from the RealEstate10K test set~\citep{zhou2018stereo}, 300 sequences from the Sekai-Real test subset~\citep{li2025sekai}, and 500 unseen sequences from the SpatialVID dataset. These samples cover diverse subjects and a wide range of camera trajectories. Reported results are averaged across all sequences.
Following ReCamMaster~\citep{bai2025recammaster}, we additionally report frame–text similarity (\textbf{CLIP-T}) and inter-frame temporal consistency (\textbf{CLIP-F})~\citep{kuang2024collaborative}, together with results on the widely used \textbf{VBench} metrics~\citep{huang2024vbench}.

\noindent \textbf{Results and Analysis.}
% camera
Quantitative results for camera-controlled video generation in ~\tabref{tab:cvd_compare} compare models trained on different datasets across different benchmarks, all evaluated with camera pose and text as inputs. The analysis reveals that the model trained on \textit{SpatialVID-HQ} consistently achieves the highest accuracy in camera control across all three benchmarks, demonstrating the high quality of the camera pose annotations and the strong generalization capability of our dataset.
% visual
Additionally, models trained on \textit{SpatialVID-HQ} attain the highest CLIP-T, reflecting enhanced visual fidelity and text-video alignment. It further shows stable improvements across all VBench metrics, particularly in Imaging Quality. Qualitative comparisons in ~\figref{fig:cvd_qualitative} show that videos generated from models trained on \textit{SpatialVID-HQ} exhibit greater realism and temporal consistency, benefiting from the dataset’s diverse trajectories, high-quality videos, and consistent text annotations.

% In conclusion, the results highlight the key advantages of our SpatialVID-HQ dataset, including its diverse trajectory types, high-quality video data, and consistent text annotations. Training on our data improves not only the model's camera controllability but also the overall visual quality of the generated videos.

%% file: sec/5_task2.tex
\subsection{Novel View Synthesis}

\begin{table*}[h!]
\caption{Comparison of Original and Fine-tuned Models for Camera Pose Estimation on Sintel~\citep{alnegheimish2022sintel}, TUM-dynamics~\citep{sturm2012benchmark}, and Dycheck~\citep{gao2022monocular}.}
\centering
\resizebox{.9\textwidth}{!}{%
\begin{tabular}{lccccccccc}
\toprule[0.17em]
% \Xhline{3\arrayrulewidth}
\multirow{2}{*}{\textbf{Method}} & \multicolumn{3}{c}{\textbf{Sintel}} & \multicolumn{3}{c}{\textbf{TUM-dynamics}} & \multicolumn{3}{c}{\textbf{Dycheck}} \\
\cmidrule(lr){2-4} \cmidrule(lr){5-7} \cmidrule(lr){8-10}
 & ATE$\downarrow$ & RPE trans$\downarrow$ & RPE rot$\downarrow$ & ATE$\downarrow$ & RPE trans$\downarrow$ & RPE rot$\downarrow$ & ATE$\downarrow$ & RPE trans$\downarrow$ & RPE rot$\downarrow$\\
\cmidrule{1-10}
CUT3R & \textbf{0.210} & 0.070 & 0.637 & 0.049 & 0.015 & 0.449 & 0.020 & \textbf{0.011} & 1.275 \\
CUT3R (Fine-tuned) & \textbf{0.210} & \textbf{0.069} & \textbf{0.619} & \textbf{0.040} & \textbf{0.013} & \textbf{0.395}  & \textbf{0.019}  & \textbf{0.011}  & \textbf{1.184} \\
\midrule
VGGT & \textbf{0.134} & 0.079 & 0.501 & 0.015 & 0.013 & 0.352  & \textbf{0.005} & \textbf{0.008} & \textbf{1.087} \\
VGGT (Fine-tuned) & 0.148 & \textbf{0.075} & \textbf{0.462} & \textbf{0.013} & \textbf{0.011} & \textbf{0.312}  &  \textbf{0.005} & 0.009 & 1.090 \\
% \Xhline{3\arrayrulewidth}
\bottomrule[0.17em]
\end{tabular}}
\label{tab:pose_finetune}
  \vspace{-2mm}
\end{table*}

\noindent \textbf{Baseline.}
Recent large reconstruction models for novel view synthesis, such as GS-LRM~\citep{zhang2024gs} and Long-LRM~\citep{ziwen2025llrm}, have demonstrated strong scene-level 3D reconstruction capabilities. We adopt GS-LRM as our baseline framework, which combines a Transformer-based architecture with 3D Gaussian rendering for high-fidelity scene recovery. For fairness, GS-LRM is trained separately on RealEstate10K and a SpatialVID-HQ subset under identical configurations and data volumes. The SpatialVID-HQ subset is curated to match the number of clips in RealEstate10K, prioritizing static scenes for consistency. Each model is finetuned on 8 A6000 GPUs for 3 days and detailed training setup is provided in Supplementary Materials.

\noindent \textbf{Metrics.}
We follow GS-LRM~\citep{zhang2024gs} and evaluate novel-view synthesis using PSNR, SSIM, and LPIPS.
Experiments are conducted on 500 DL3DV~\citep{ling2024dl3dv} and 500 unseen SpatialVID sequences, with 2 reference and 4 target views per sequence, rendered at $360\times640$ for a fair comparison.

\noindent \textbf{Results and Analysis.}
As shown in~\tabref{tab:lrm_compare}, the model trained on the SpatialVID subset consistently outperforms RealEstate10K across all the metrics. These results highlight the strong generalization capability of SpatialVID, with superior performance even on the predominantly outdoor DL3DV scenes. This suggests that the diversity and quality of SpatialVID data lead to more robust scene understanding and higher-quality novel view synthesis. Further results are provided in Supplementary Materials.

\begin{table}[t]
\caption{Comparison of GS-LRM across different training datasets on DL3DV~\citep{ling2024dl3dv} and our SpatialVID.} %at 360$\times$640 resolution for fairness.}
\centering
\resizebox{1.\linewidth}{!}{%
\begin{tabular}{lcccccc}
\toprule[1.2pt]
\multirow{2}{*}{\makecell{\textbf{Training}\\\textbf{Data}}} & \multicolumn{3}{c}{\textbf{DL3DV}} & \multicolumn{3}{c}{\textbf{SpatialVID}} \\
\cmidrule(lr){2-4} \cmidrule(lr){5-7}
 & PSNR $\uparrow$ & SSIM $\uparrow$ & LPIPS $\downarrow$ & PSNR $\uparrow$ & SSIM $\uparrow$ & LPIPS $\downarrow$ \\
\midrule
RE10K & 27.01 & 0.889 & 0.132 & 24.13 & 0.774 & 0.222 \\
SpatialVID & \textbf{27.80} & \textbf{0.892} & \textbf{0.116} & \textbf{24.97} & \textbf{0.790} & \textbf{0.203} \\
\toprule[1.2pt]
\end{tabular}}
\label{tab:lrm_compare}
  \vspace{-2mm}
\end{table}

%% file: sec/5_task3.tex
\subsection{Geometric Prediction}
% VGGT/MonST3R/CUT3R
% \begin{table*}[t]
% \centering
% \resizebox{\linewidth}{!}{%
% \begin{tabular}{lccccccccc}
% \Xhline{3\arrayrulewidth}
% \multirow{2}{*}{\textbf{Method}} & \multicolumn{3}{c}{\textbf{Sintel}} & \multicolumn{3}{c}{\textbf{TUM-dynamics}} & \multicolumn{3}{c}{\textbf{ScanNet}} \\
% \cmidrule(lr){2-4} \cmidrule(lr){5-7} \cmidrule(lr){8-10}
%  & ATE$\downarrow$ & RPE trans$\downarrow$ & RPE rot$\downarrow$ & ATE$\downarrow$ & RPE trans$\downarrow$ & RPE rot$\downarrow$ & ATE$\downarrow$ & RPE trans$\downarrow$ & RPE rot$\downarrow$ \\
% \cmidrule{1-10}
% CUT3R & 0.210 & 0.071 & \textbf{0.627} & 0.045 & 0.014 & 0.441 & 0.096 & 0.022 & 0.733 \\
% CUT3R* & \textbf{0.178} & \textbf{0.055} & 0.651 & \textbf{0.041} & \textbf{0.013} & \textbf{0.374} & \textbf{0.095} & 0.022 & \textbf{0.604} \\
% \midrule

% MonST3R & 0.210 & 0.071 & \textbf{0.627} & 0.045 & 0.014 & 0.441 & 0.096 & 0.022 & 0.733 \\
% MonST3R* & \textbf{0.178} & \textbf{0.055} & 0.651 & \textbf{0.041} & \textbf{0.013} & \textbf{0.374} & \textbf{0.095} & 0.022 & \textbf{0.604} \\
% \Xhline{3\arrayrulewidth}
% \end{tabular}%
% }
% \caption{Comparison of Original and Fine-tuned Models for Camera Pose Estimation on Sintel\wjh{cite} TUM-dynamics\wjh{cite}. }
% \label{tab:pose_finetune}
% \end{table*}

\noindent \textbf{Baselines.}
For camera pose estimation task, we adopt CUT3R~\citep{cut3r} and VGGT~\citep{wang2025vggt} as representative baselines, both demonstrating strong performance in geometric and structural reconstruction. Each model is initialized from its official pre-trained weights and fine-tuned under comparable settings. CUT3R is further fine-tuned on SpatialVID to assess pose consistency improvements, while VGGT is trained on its original data both with and without SpatialVID-HQ for fair comparison. Each model is fine-tuned on 8 H20 GPUs for 2 days and detailed training setup is provided in Supplementary Materials.

\noindent \textbf{Metrics.}
For geometric prediction we focus on camera-pose estimation in dynamic scenes. We evaluate models on three diverse benchmarks including Sintel~\citep{alnegheimish2022sintel}, TUM-dynamics~\citep{sturm2012benchmark}, and Dycheck~\citep{gao2022monocular}, all containing non-rigid and dynamic objects to ensure comprehensive coverage of different data types and scene characteristics. Following previous works ~\citep{zhang2024monst3r,cut3r}, we report ATE, RPE-rot and RPE-trans computed over estimated camera trajectories.

% Both CUT3R and VGGT models are trained on large-scale datasets with 3D annotations and 已经有了很好的效果, 训练数据中 most of which are mainly synthetic data. SpatialVID spanning diverse real-world scenes can help stimulating potential 增强泛化性 in some realistic cases. according to \tabref{tab:pose_finetune}, both models achieves significant improvements on the TUM-dynamics benchmark, wich is a 真实世界benchmark. On the synthetic Sintel benchmark, 模型均有一定程度的提升，仅VGGT在ATE指标有一点变差。The Dycheck benchmark, characterized by high dynamics and complex camera trajectories caused by handheld jitter, poses substantial reconstruction challenges. After fine-tuning, CUT3R’s metrics remain stable or slightly decrease. As a more powerful model, VGGT already delivers near-optimal predictions, making slight increases in some metrics post-fine-tuning reasonable.

\noindent \textbf{Results and Analysis.}
Both CUT3R and VGGT are trained on multiple 3D-annotated datasets and already achieve strong results, though most training data are synthetic. SpatialVID, covering diverse real-world scenes, complements these sources and improves generalization to realistic conditions. As shown in \tabref{tab:pose_finetune}, fine-tuning on SpatialVID yields notable gains for both models on the TUM-dynamics benchmark. On the synthetic Sintel benchmark, both models show improvements, except for a slight ATE regression for VGGT. The Dycheck benchmark, with high dynamics and complex camera trajectories from handheld jitter, remains challenging. After fine-tuning, CUT3R’s performance is marginally improved, while VGGT, near its ceiling, exhibits only minor fluctuations, reflecting its strong pre-trained capability.

%% file: sec/7_conclusion.tex
\section{Discussion and Conclusion}
\paragraph{Limitations.}
Our annotation pipeline inherits failure modes from MegaSaM and can degrade under extreme scenarios (e.g., object-dominated frames, varying focal lengths, and severe radial distortion), constraining its broader application. Additionally, the predicted camera poses exhibit non-metric properties in specific scenarios, and masks derived from motion probabilities yield suboptimal performance in complex scenes. 
We expect that the development of advanced video pose estimator (e.g., ViPE~\citep{huang2025vipe}) would mitigate these issues in the future.

\paragraph{Conclusion.}
We introduce \DATAname, a large-scale video dataset spanning diverse real-world scenes, with tightly aligned semantic and geometric annotations. Our procedural pipeline distills in-the-wild videos into clips annotated with camera motion, depth, and structured, motion-aware scene descriptions. By unifying explicit 3D motion control with rich textual semantics, \DATAname provides strong 3D inductive biases for physically grounded video generation, dynamic scene synthesis, and spatial intelligence. Extensive experiments demonstrate its effectiveness across multiple tasks, establishing a robust foundation for future research.

%% file: sec/suppl.tex
\maketitlesupplementary
\appendix
% data flow of pipeline
\begin{figure*}[t] %[t!]
    \centering
    \begin{overpic}[width=1.\textwidth,
    % \begin{overpic}[width=1.\columnwidth,
    % height=8cm,
    trim=0 0 60 0,
    clip
    ]{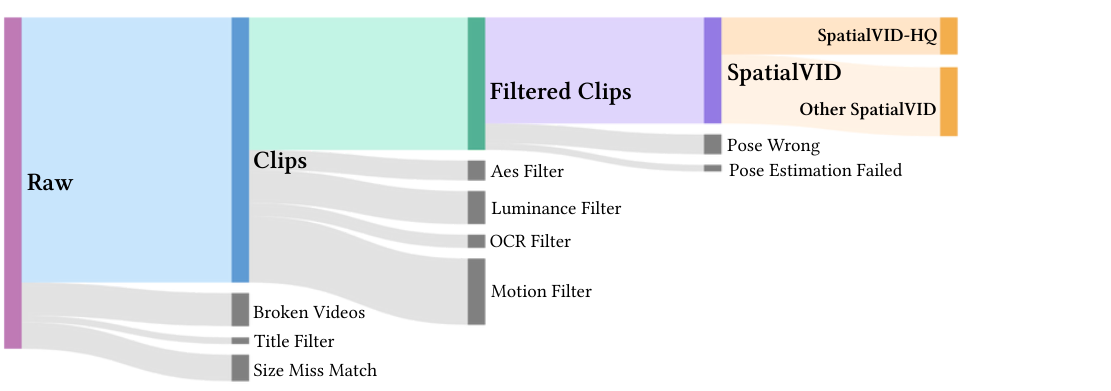}
    \end{overpic}
  \caption{
    \textbf{The data flow of video filtering}.
    Raw videos are first pre-filtered to exclude content with quality defects, incorrect dimensions, or irrelevant titles. The remaining videos are segmented into clips, which are then ranked via a hierarchical scoring strategy integrating aesthetics metrics, luminance, OCR, and motion values. High-scoring clips undergo a dual annotation pipeline to capture both spatial structure and semantic information, yielding the final \DATAname~dataset. This pipeline is also employed to curate a high-quality subset (\DATAname-HQ) with a more balanced category distribution.
    }
% spatial coordinates -> spatial structure
  \label{fig:data_flow}
\end{figure*} 

\section*{Overview}

This supplementary material provides extended details and additional results complementing the main paper. We elaborate on the data curation pipeline (\secref{sec:supp:curation}), present in-depth analyses of dataset statistics and semantic properties (\secref{sec:supp:analysis}), and describe additional validation tasks and implementation details (\secref{sec:supp:validation}). An overview of the video filtering and annotation process is shown in \figref{fig:data_flow}.

\section{Details of Our Curation Pipeline}
\label{sec:supp:curation}
\begin{figure*}[t]
    \centering
    \begin{overpic}[width=1.\textwidth,
    clip
    ]{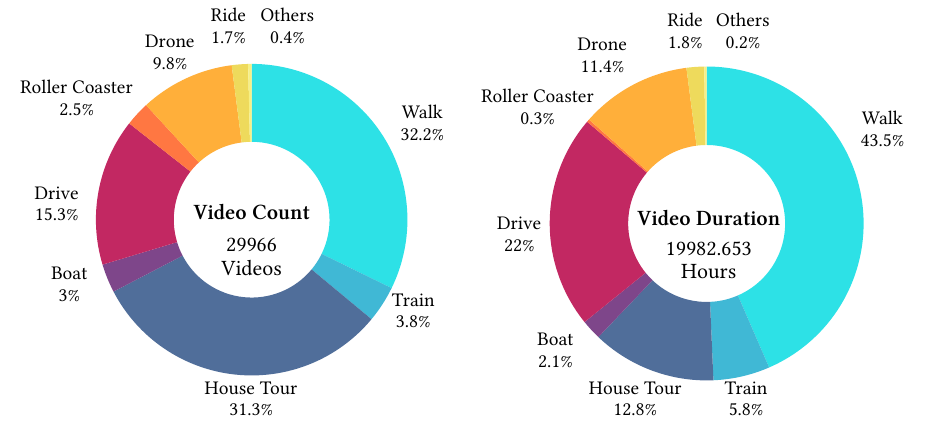}
    \end{overpic}
  \caption{
  \textbf{Statistics of pre-filtered videos} (\textit{aka}. the \enquote{Clips} in the \figref{fig:data_flow}).
      The left panel shows the quantity distribution of raw videos, while the right panel presents the duration distribution. These charts illustrate the variety of shooting contexts, including indoor (house tour) and outdoor (walking, train, drone, etc.) scenarios, demonstrating broad coverage of different shooting carriers and environments.}
  \label{fig:raw_video_dist}
\end{figure*}

\figref{fig:data_flow} presents the overall data flow of our video curation process, through which the raw videos are progressively refined into the final \DATAname~dataset. The pipeline integrates automatic quality filtering, geometric annotation, and semantic captioning. Figure~\ref{fig:raw_video_dist} further illustrates the duration and quantity distribution of the collected raw data, covering diverse indoor and outdoor scenes such as walking, train rides, and drone flights.

% raw video distribution illustration

\subsection{Score Filtering}
To ensure data quality, we apply four complementary filters based on \textit{aesthetics}, \textit{luminance}, \textit{text content}, and \textit{motion intensity}. These filters remove visually poor or unsuitable clips before any geometric or semantic processing, significantly improving the robustness of downstream pose estimation.

\begin{figure*}[h!]
    \centering
    \begin{subfigure}{1.\linewidth}
        \centering
        \includegraphics[width=1.\textwidth]{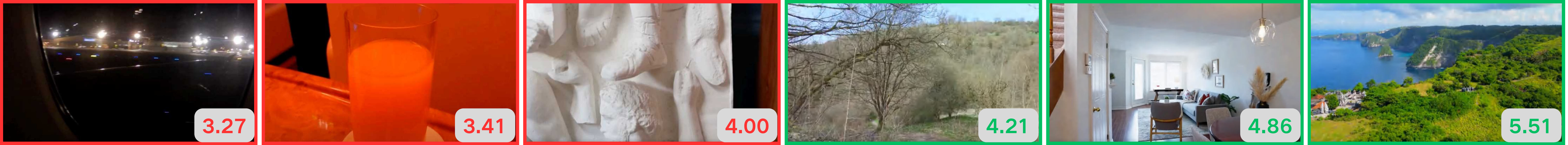} 
        \vspace{-5mm}
        \caption{Aesthetics Filtering}
        \label{fig:aes}  
    \end{subfigure}%
    \hfill
    \begin{subfigure}{1.\linewidth}
        \centering
        \includegraphics[width=1.\textwidth]{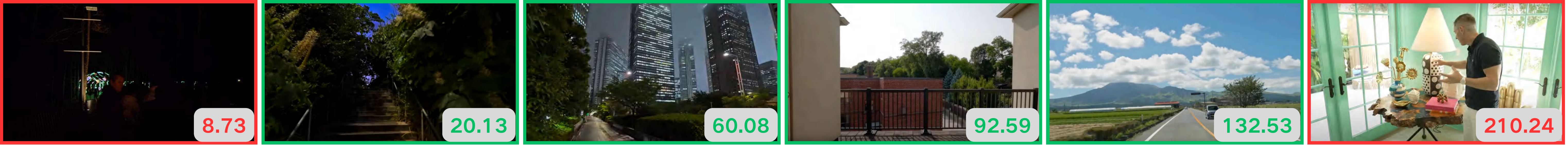}  
        \vspace{-5mm}
        \caption{Luminance Filtering} 
        \label{fig:lum}  
    \end{subfigure}%
    \hfill
    \begin{subfigure}{1.\linewidth}
        \centering
        \includegraphics[width=1.\textwidth]{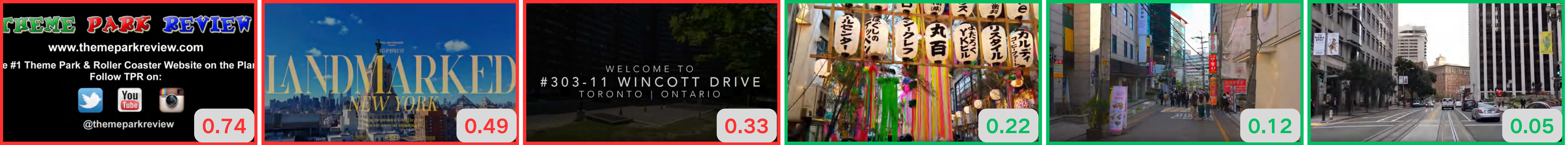}  
        \vspace{-5mm}
        \caption{OCR Filtering}  
        \label{fig:ocr}  
    \end{subfigure}
    \hfill
    \begin{subfigure}{1.\linewidth}
        \centering
        \includegraphics[width=1.\textwidth]{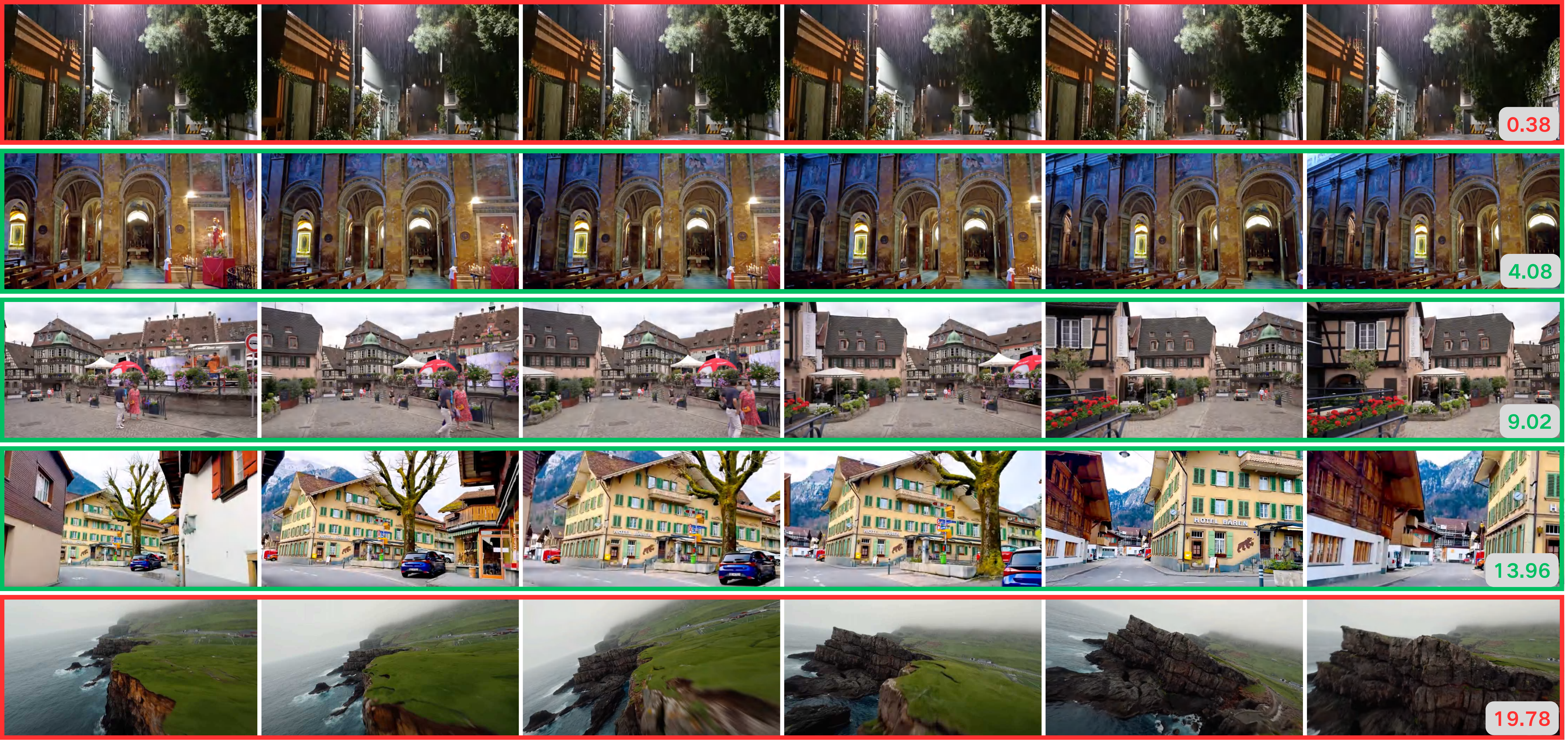}  
        \caption{Motion Filtering}  
        \label{fig:motion}  
    \end{subfigure}%
    \caption{\textbf{Video filtering strategies}. Videos are filtered based on various quality criteria (\textit{Aesthetics}, \textit{Luminance}, \textit{OCR}, and \textit{Motion}). The number in the bottom-right corner of each clip represents its score for the corresponding quality filter. Clips with green boxes are retained, while those with red boxes are discarded due to scores below the threshold.}
    \label{fig:score_filter}
\end{figure*}

\textbf{Aesthetic Filtering}. To quantitatively assess visual appeal, we use a CLIP + MLP aesthetic score predictor~\citep{schuhmann2022laion}. The model assigns a score from 0 to 10, with higher values indicating better quality. For each video clip, the score is averaged across the first, middle, and last frames. Clips with an average score below 4.0 are considered insufficiently appealing and discarded \figref{fig:aes}).

\textbf{Luminance Filtering}. Luminance is calculated for the first, middle, and last frames using the standard formula \(L = 0.2126 R + 0.7152 G + 0.0722 B\), where \(R\), \(G\), and \(B\) are the respective channel values. Clips with average luminance outside the range [20, 140], either too dark or too bright, are excluded, ensuring that only clips with proper exposure are retained (\figref{fig:lum}).

\textbf{OCR filter}. For text detection, we use the latest release of PaddleOCR~\citep{cui2025paddleocr}, which offers high accuracy and robust multilingual support. We processe the first, middle, and last frames of each clip to detect text regions, computing the ratio of text area to frame size. Clips where the text area exceeded 30\% are removed, as these are considered informational rather than visual (\figref{fig:ocr}).

\textbf{Motion Filtering}. We use lightweight VMAF~\citep{VMAF2016}, which is integrated into FFmpeg and outputs a valid motion score ranging from 2.0 to 14.0 (\figref{fig:motion}).

% score filtering fig

\subsection{Geometry Pipeline}

\begin{figure*}[t]
    \centering
    \begin{overpic}[width=1.\textwidth,
    clip
    ]{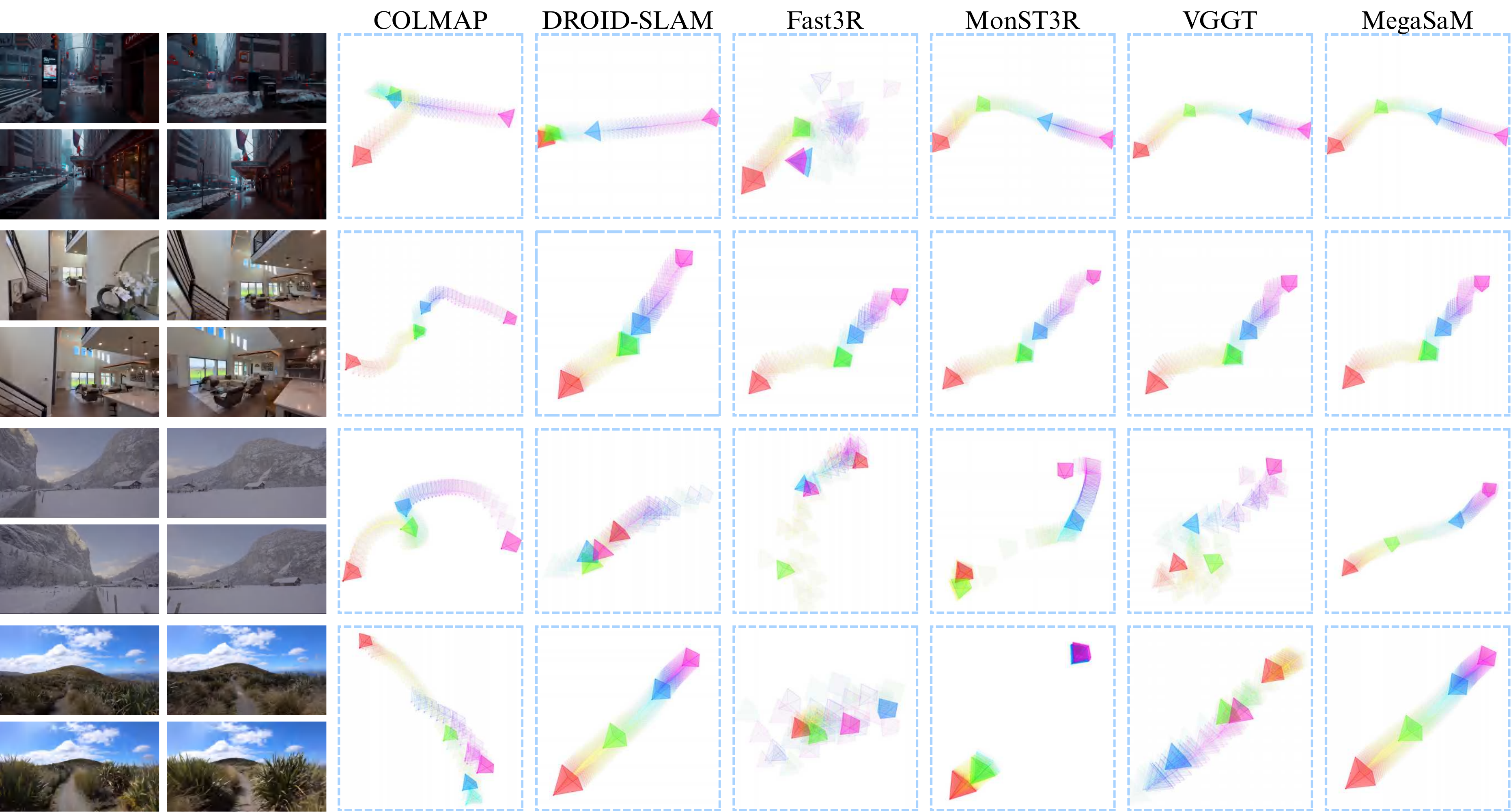}
    \end{overpic}
  \caption{\textbf{Comparison of MegaSaM with other SLAM/3D reconstruction methods}. We visualize the trajectories predicted by six representative methods. The color order \textcolor{red}{R}\textcolor{orange}{O}\textcolor{yellow}{Y}\textcolor{green}{G}\textcolor{blue}{B}\textcolor{violet}{V} corresponds to the progression from the initial to the final time step.}
  \label{fig:compare_megasam}
  % \vspace{-5mm}
\end{figure*}

% wjh revise 1108
We employ \textbf{MegaSaM} as our primary geometric reconstruction engine. As shown in \figref{fig:compare_megasam}, it achieves superior accuracy and robustness compared with DROID-SLAM~\citep{teed2021droid}, COLMAP~\citep{schoenberger2016sfm}, and Fast3R~\citep{yang2025fast3r}, while being faster than MonST3R~\citep{zhang2024monst3r}. Unlike VGGT~\citep{wang2025vggt}, MegaSaM also maintains stability in feature-sparse scenes. We utilize three trajectory statistics for quality assessment:

\textbf{Move Distance (MoveDist)}.
Total camera travel distance, computed as the sum of Euclidean distances between consecutive camera centers.

\textbf{Rotation Angle (RotAngle)}.
Cumulative angular displacement across frames, capturing the extent of viewpoint change.

\textbf{Trajectory Turns (TrajTurns)}.
Number of significant turns, estimated by counting local extrema in the sequence of orientation angles relative to a start–end reference direction.

% figure about method compare

\subsection{Caption Pipeline}
The captioning process integrates visual-language reasoning and structured text generation in two stages.

\begin{figure*}[t!]
    \centering
    \begin{overpic}[width=1.\textwidth,
    clip
    ]{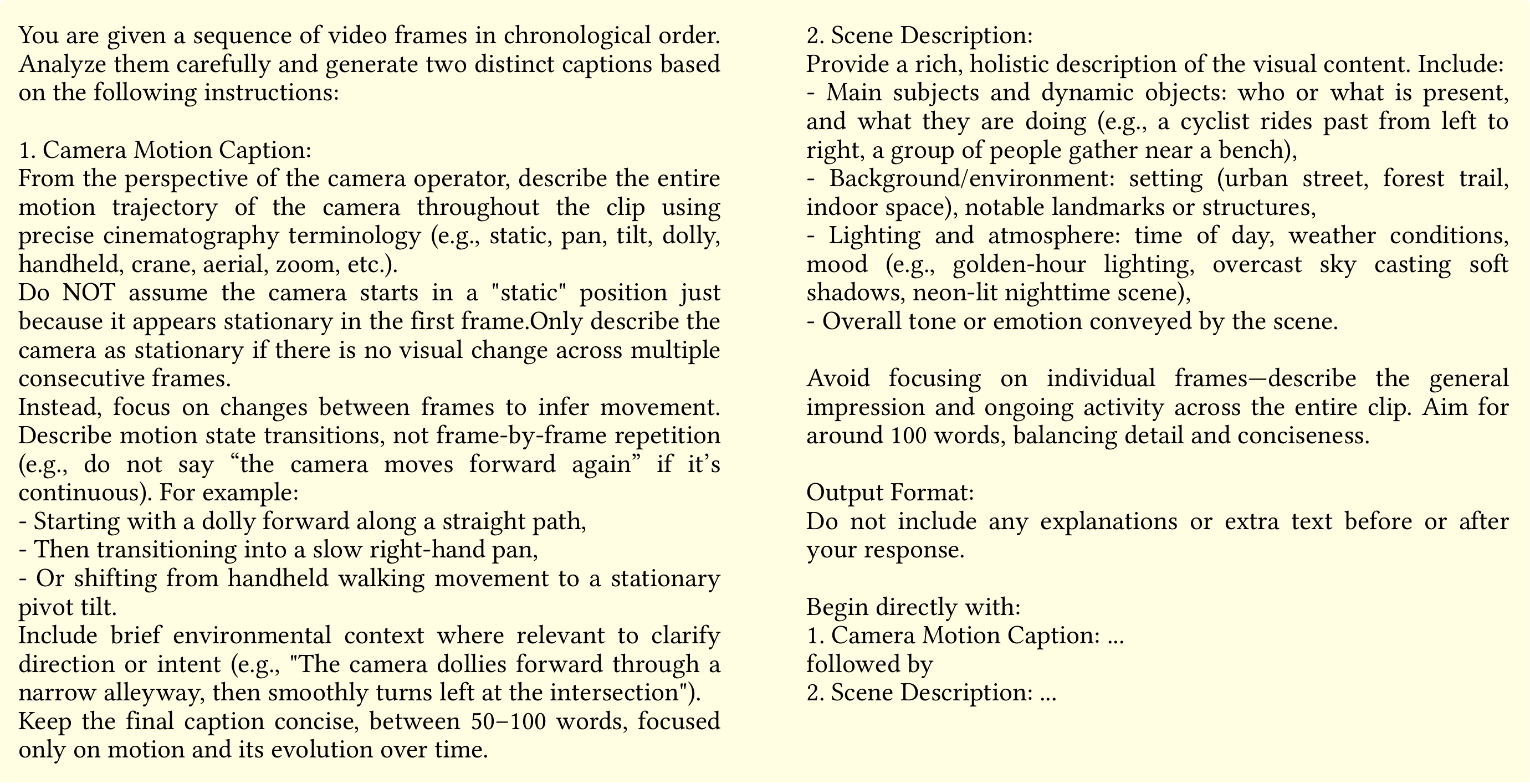}
    \end{overpic}
    \caption{\textbf{Visual-language model (VLM) prompt}.
    The template guides Gemini-2.0-flash to describe both the visual content and coarse camera motion in natural language, enabling structured parsing of dynamic scenes.
    }
  \label{fig:vlm_prompt}
\end{figure*}

\textbf{Stage 1: Visual Parsing}.
We use Gemini-2.0-flash~\citep{team2023gemini} to analyze sampled frames (1 fps), producing an initial description of camera motion and a summary of scene layout. The prompting format is illustrated in \figref{fig:vlm_prompt}.

\begin{figure*}[t!]
    \centering
    \begin{overpic}[width=1.\textwidth,
    % trim=20 100 20 50, 
    clip
    ]{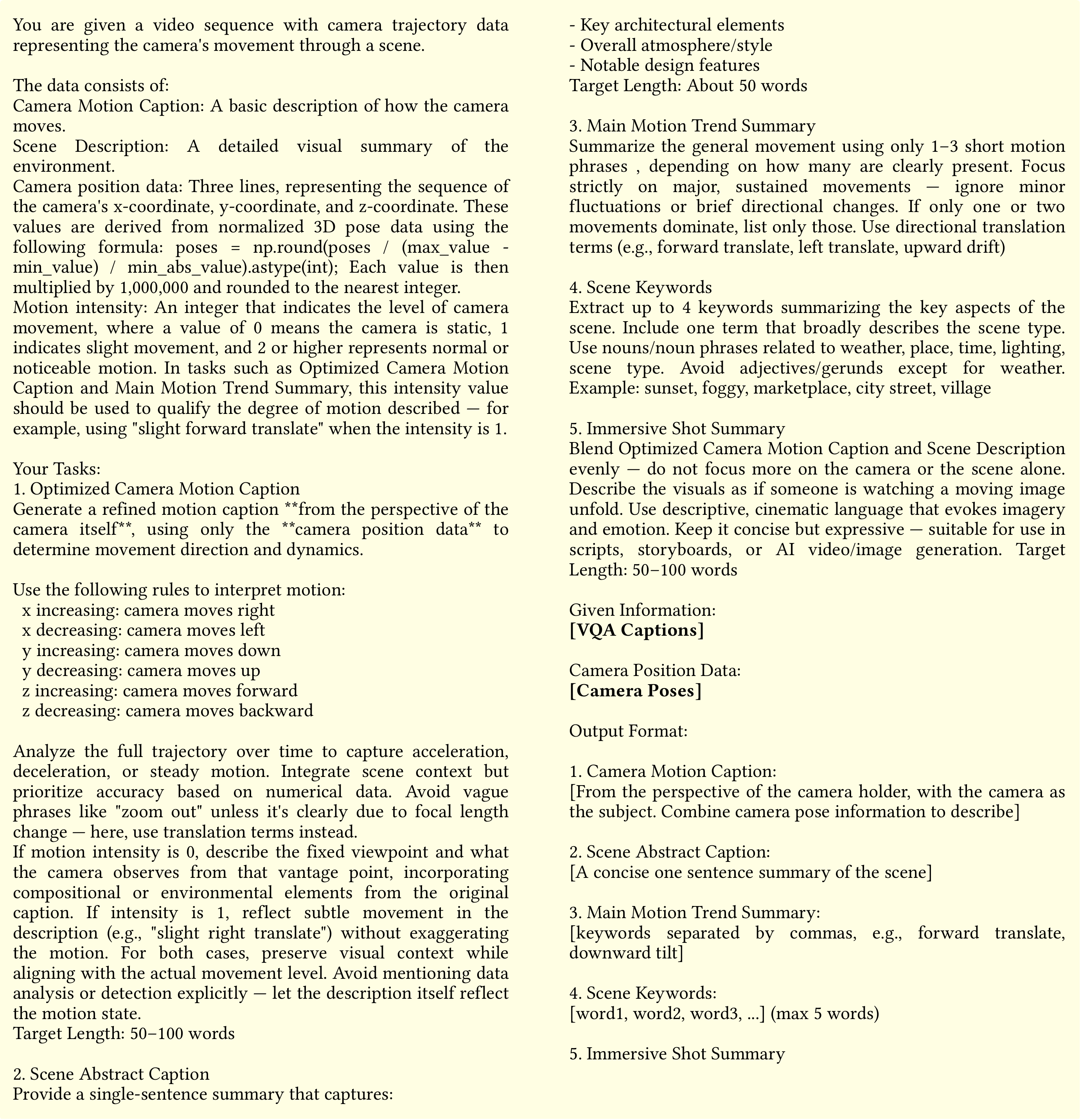}
    \end{overpic}
    \caption{\textbf{Large-language model (LLM) refinement prompt}.
    This instruction template conditions Qwen3-30B-A3B to generate coherent, attribute-rich captions aligned with the extracted spatial and motion cues.
    }
  \label{fig:llm_prompt}
\end{figure*}

\textbf{Stage 2: Language Refinement}.
The outputs, along with calibrated camera poses, are then refined using Qwen3-30B-A3B~\citep{yang2025qwen3}. This stage yields (1) concise scene abstracts, (2) immersive shot-level narratives, and (3) structured semantic tags describing scene type, lighting, weather, crowd density, and time of day. Additionally, \textit{Motion Trends} labels (e.g., pan, dolly, rotate, steady) summarize camera dynamics. Distributions of these tags are shown in \figref{fig:motion_dist}. The prompt template used for the large-language refinement is shown in \figref{fig:llm_prompt}.

% need figref in paragraphs
% figure about VLM prompt

% figure about LLM prompt

\subsection{Instruction Examples}
\begin{figure*}[t]
\centering
\begin{overpic}[width=1.\textwidth, clip]{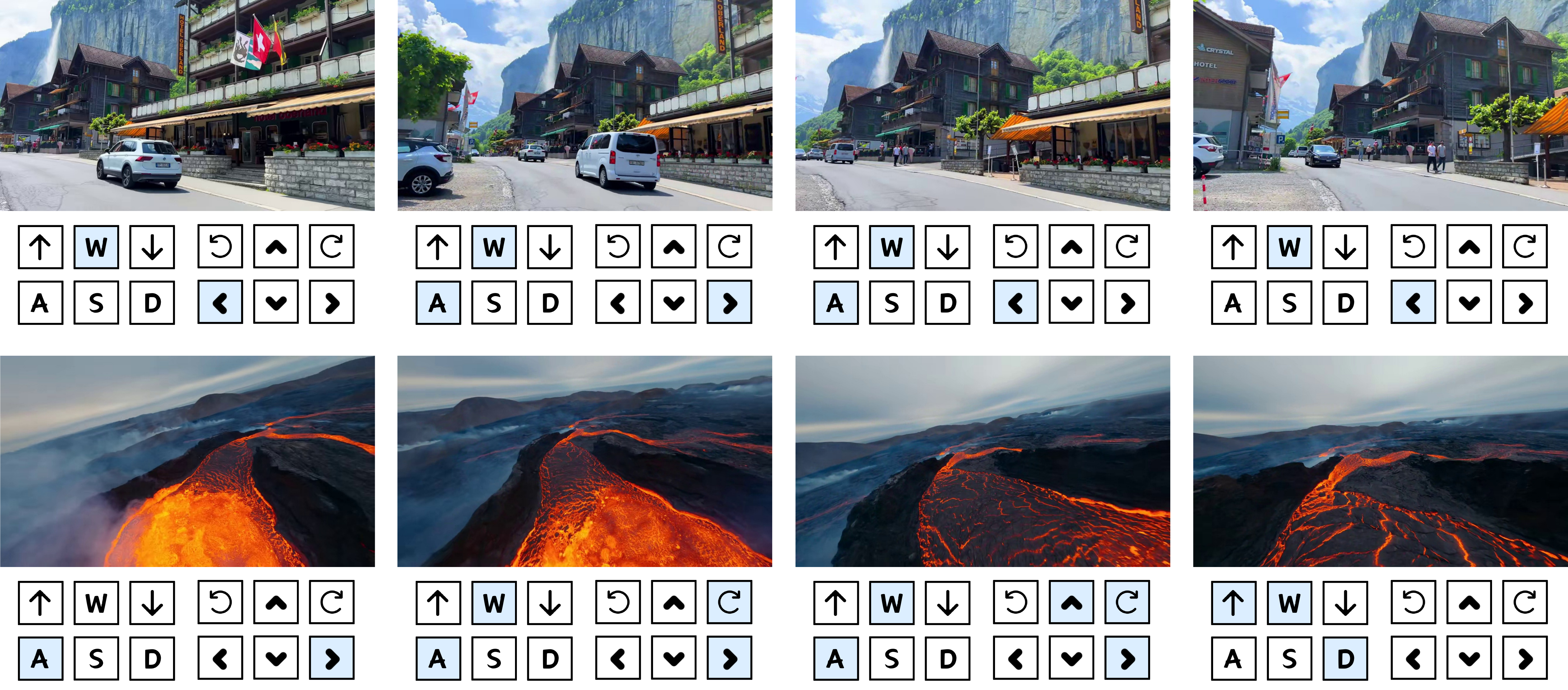}
\end{overpic}
\caption{
\textbf{Examples of motion instructions.}
Keyboard-style icons denote camera motion directions. The cluster on the left corresponds to translations: \textbf{W} and \textbf{S} indicate forward and backward movement, \textbf{A} and \textbf{D} indicate left and right movement, and $\uparrow,\downarrow$ represent vertical movement. The cluster on the right corresponds to rotations: arrows denote pitch ($\wedge,\vee$) and yaw ($<,>$), while circular arrows indicate roll ($\circlearrowleft,\circlearrowright$). These visual cues intuitively describe camera operations, linking geometric motion to semantic labels.
}
% \vspace{-6mm}
\label{fig:instruction_vis}
\end{figure*}

Examples of motion instructions are illustrated in \figref{fig:instruction_vis}. Each instruction corresponds to a specific type of camera movement derived from pose dynamics, providing an interpretable bridge between geometric motion and textual representation. The visualizations demonstrate how camera translations and rotations are systematically mapped to human-readable motion terms, ensuring clarity and consistency across clips.

\section{Details of Dataset Analysis}
\label{sec:supp:analysis}
This section provides extended analyses of the \DATAname~dataset, focusing on semantic composition, caption statistics, and qualitative examples. We examine the distributions of camera motion and scene attributes, analyze caption diversity, and visualize representative samples from the dataset.

\subsection{Semantic Analysis}
\begin{figure*}[t!]
    \centering
    \begin{overpic}[width=1.\textwidth,
    % trim=20 100 20 50, 
    clip
    ]{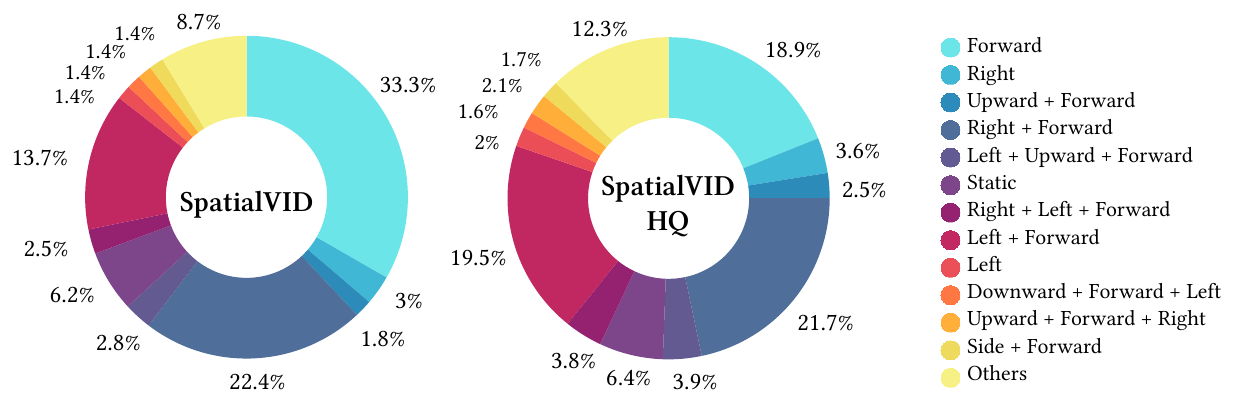}
    \end{overpic}
    \caption{\textbf{Distribution of camera motion directions}. The donut charts show the distribution of camera motion directions for the \DATAname~(left) and HQ \DATAname~(right) datasets. The original \DATAname~dataset exhibits a wide range of motion patterns. In contrast, the HQ \DATAname~dataset features a more balanced distribution, addressing the overrepresentation of any single motion direction.}
  \label{fig:motion_dist}
\end{figure*}

\textbf{Camera Motion Distribution}.
\figref{fig:motion_dist} shows the distribution of camera motion directions across \DATAname~and its high-quality subset, \DATAname-HQ. The original dataset contains a wide range of motion types, including forward, lateral, and rotational movements, but their distribution is not well-balanced. In contrast, \DATAname-HQ displays a more balanced distribution, mitigating bias toward any single motion direction and offering improved diversity for motion-conditioned generation or control tasks.

\begin{figure*}[t!]
    \centering
    \begin{subfigure}{0.5\textwidth}
        \centering
        \includegraphics[width=1.\linewidth]{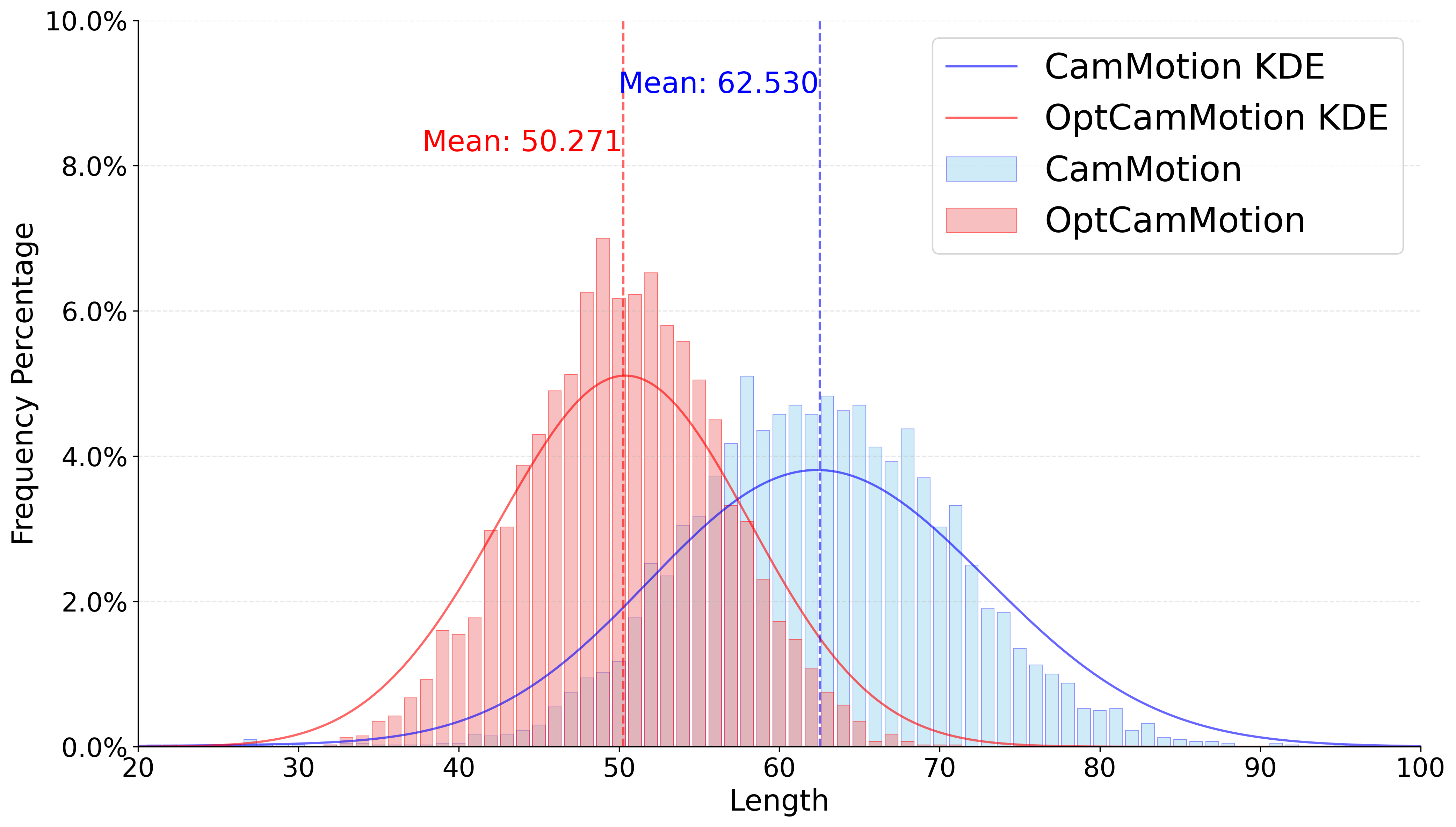} 
        \caption{Motion caption length distribution}
        \label{fig:motion_caption_len_distribution}  
    \end{subfigure}%
    \hfill
    \begin{subfigure}{0.5\textwidth}
        \centering
        \includegraphics[width=1.\linewidth]{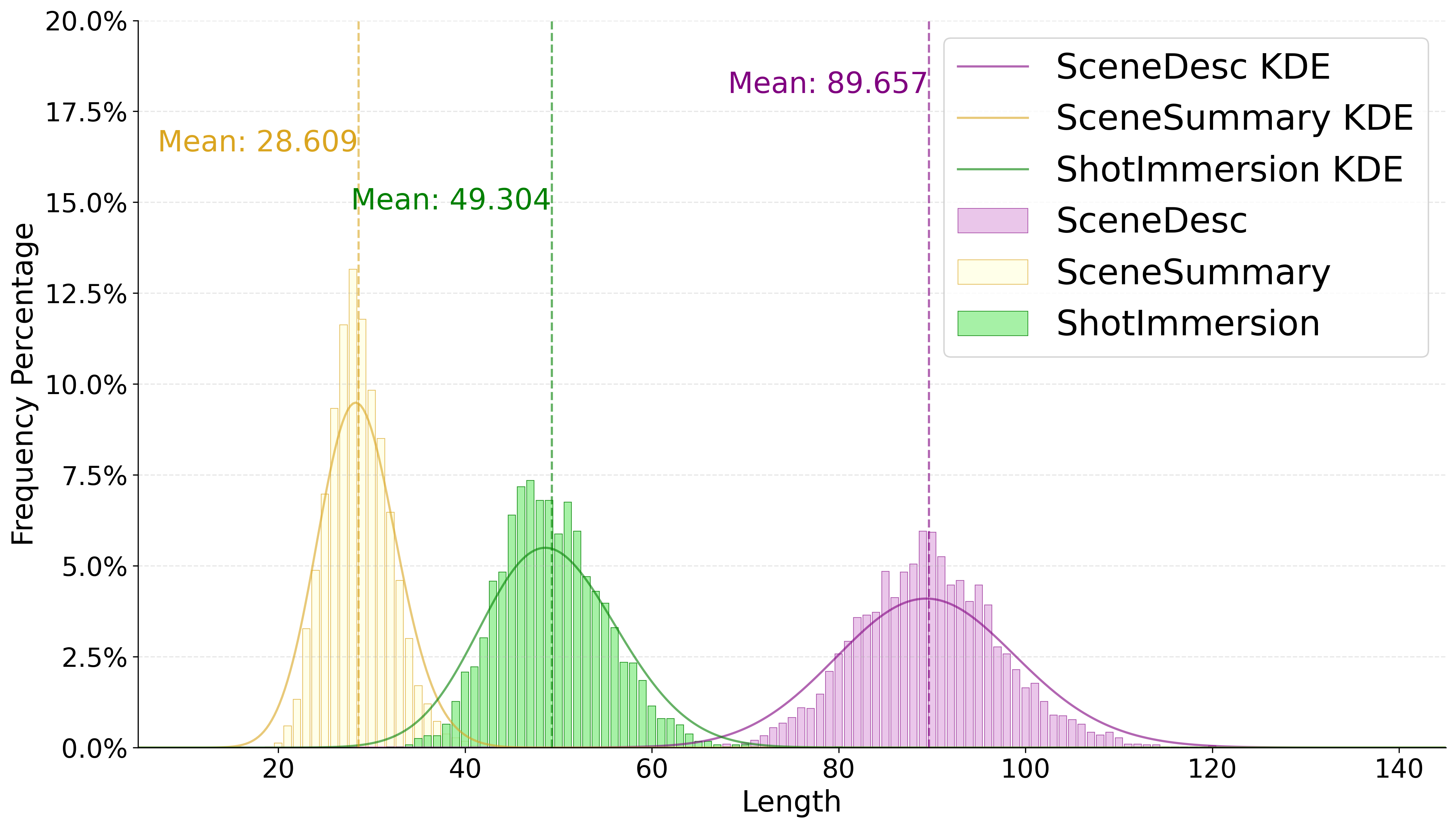}  
        \caption{Scene caption length distribution}  
        \label{fig:scene_caption_len_distribution}  
    \end{subfigure}%
    \hfill
\caption{
    \textbf{Statistical analysis of the caption data}. 
     Fig. (a) and Fig. (b) show the length distributions for motion and scene captions, respectively, comparing the original captions to our enhanced versions. A significant increase in caption length is evident for both types after enhancement. 
}
    \label{fig:caption_analysis}
\end{figure*}

\textbf{Caption Length and Enrichment}.
We provide multi-level captions for each video, including motion-oriented descriptions, concise scene summaries, and immersive narratives. To evaluate caption quality, we compare the length distributions of original and enhanced captions (\figref{fig:caption_analysis}). Both motion and scene captions show notable length increases after refinement, reflecting richer context and more detailed spatial reasoning introduced by our LLM-based generation process.

\begin{figure*}[t!] %htbp
    \centering % Center the entire figure content
    \setlength{\fboxsep}{0pt}
    \setlength{\fboxrule}{0pt}
    \newlength{\figheight}
    \setlength{\figheight}{10cm}
    % Subfigure for the left image (Donut charts)
    \begin{subfigure}[b]{0.52\textwidth}
        \centering
        \begin{overpic}[height=\figheight, width=\linewidth, keepaspectratio]{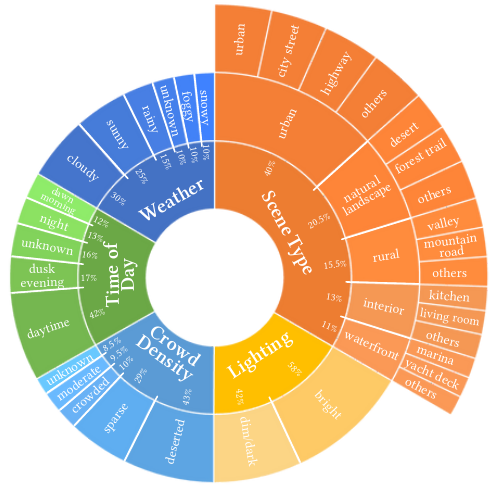}
        \end{overpic}
        \caption{Scene tags distribution}
        \label{fig:scene_tags_dist}
    \end{subfigure}
    \hfill
    \begin{subfigure}[b]{0.46\textwidth}
        \centering
        \includegraphics[height=\figheight, width=\linewidth, keepaspectratio]{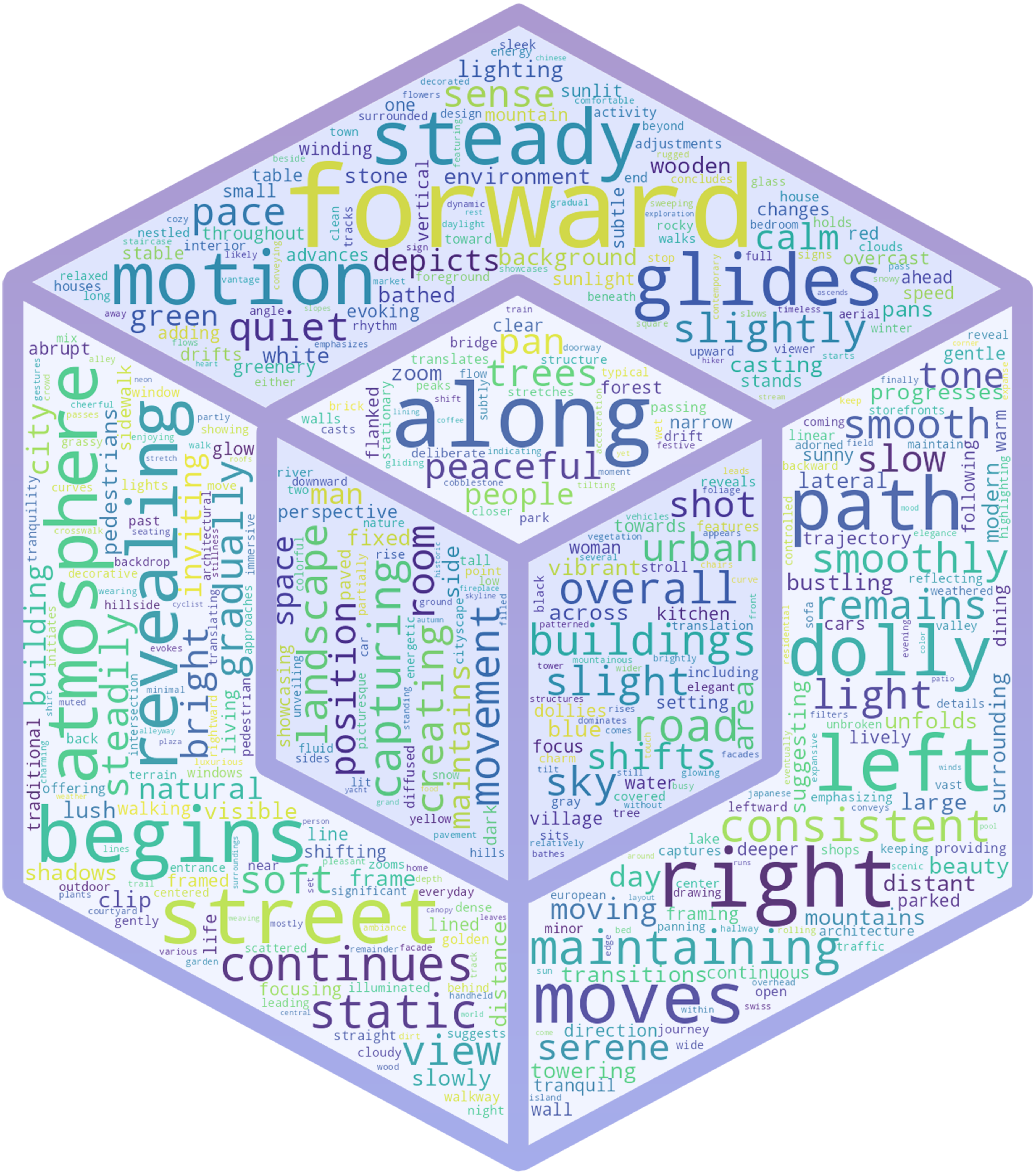}
        \caption{World cloud}
        \label{fig:wordcloud_icon}
    \end{subfigure}

    % % Main caption for the entire figure
    \caption{
    \textbf{(a) Distribution of scene tags}. The sunburst chart shows the distribution of categorical tags across five primary attributes: weather, time of day, crowd density, lighting, and scene type. The \textit{scene type} attribute is hierarchical, with sub-categories for more detailed classification. The width of each sector reflects the prevalence of the corresponding tag in the dataset.
    \textbf{(b) Word cloud}. A word cloud shaped into the \DATAname~logo, generated from the enhanced captions. The size of each word corresponds to its frequency in the corpus. Key terms such as \textit{motion}, \textit{forward}, \textit{left}, and \textit{right} emphasize the dataset's focus on describing camera movement and spatial dynamics.
    }

    \label{fig:cloud_n_dist} % A new label for the combined figure
\end{figure*}

\textbf{Hierarchical Scene Tags}.
\figref{fig:cloud_n_dist} visualizes the structured semantic attributes extracted from enhanced captions. The sunburst chart summarizes the distribution of five primary attributes—weather, time of day, crowd density, lighting, and scene type. The hierarchical scene-type branch covers fine-grained subcategories such as \textit{street}, \textit{park}, \textit{interior}, and \textit{vehicle}. The accompanying word cloud, shaped into the \DATAname~logo, highlights the dataset’s emphasis on spatial and motion-oriented vocabulary, with frequently occurring terms like \textit{motion}, \textit{forward}, and \textit{left}.

\textbf{Multi-Level Caption Design}.
SpatialVID delivers a versatile caption suite suitable for various research needs:
(1)~\texttt{OptCamMotion} provides concise, machine-friendly kinematic instructions, reducing average caption length from 62.5 to 50.3 words for clean motion supervision.
(2)~\texttt{SceneSummary} offers compact high-level context with an average of 28.6 words.
(3)~\texttt{ShotImmersion} integrates scene semantics and camera motion into rich narratives averaging 89.7 words, supporting reasoning-intensive tasks such as video understanding and story grounding.

Overall, this structured annotation design ensures both interpretability and flexibility, enabling downstream applications ranging from camera control to multimodal spatial reasoning.

% motion caption analysis

% caption length analysis

% tags and worldcloud

\subsection{Examples of SpatialVID}
\label{sec:supp:example}

\begin{figure*}[htbp]
    \centering
    \begin{overpic}[width=.95\textwidth,
    clip
    ]{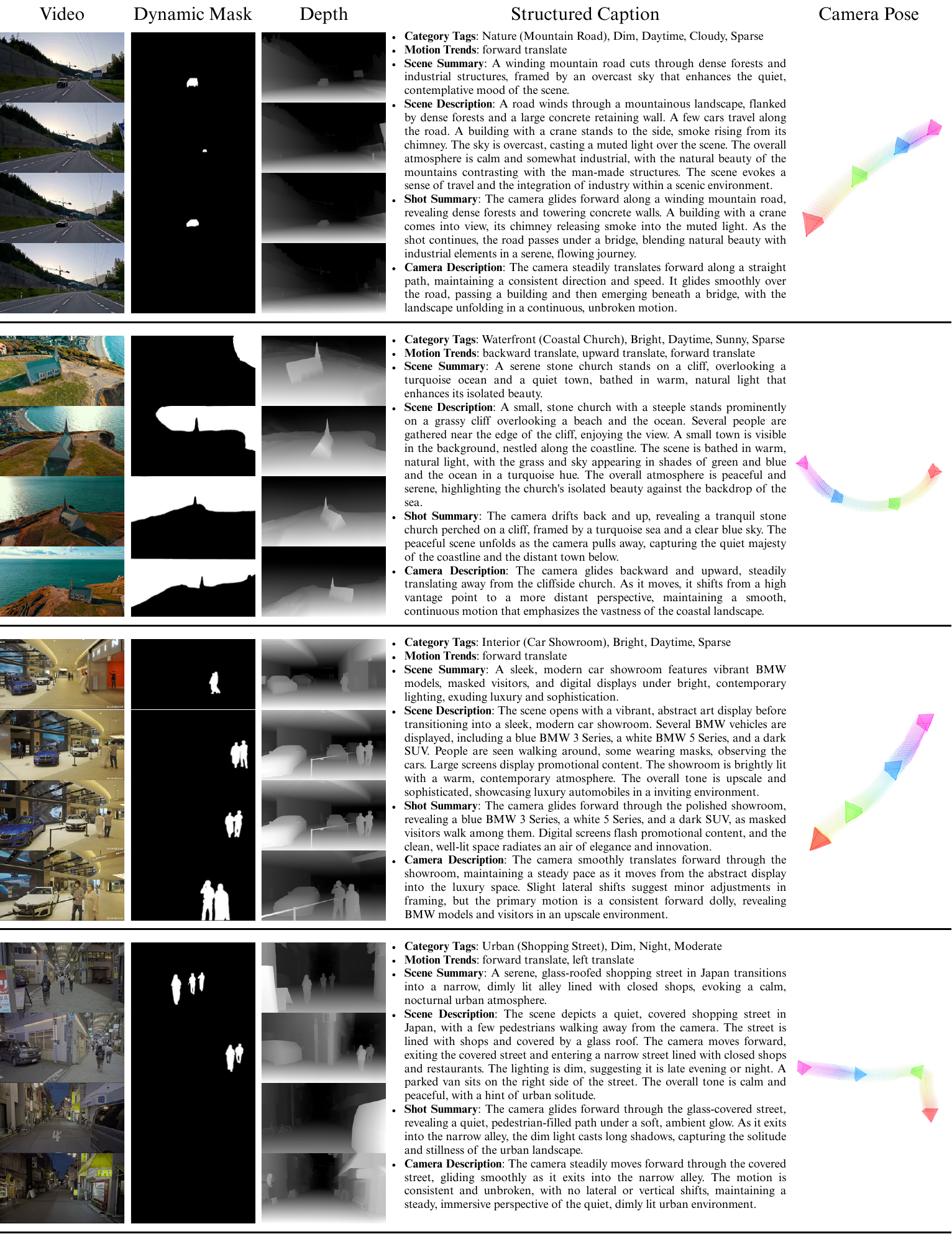}
    % \vspace{-8mm}
    \end{overpic}
  \caption{\textbf{Sample videos from \DATAname}.
    Each example includes synchronized geometry, captions, and spatial annotations. The dataset encompasses diverse environments and camera motions, highlighting its broad coverage and multimodal consistency.
    }
  \label{fig:vid_sample}
\end{figure*}

We present qualitative examples in \figref{fig:vid_sample}. The selected clips illustrate diverse motion trajectories, scene contexts, and annotation richness. Each sample contains synchronized geometry, caption, and metadata, demonstrating the spatial and semantic consistency maintained throughout the dataset.

% qualitative
\begin{figure*}
    \centering
    \vspace{-4mm}
    \begin{overpic}[width=.95\textwidth,
    clip
    ]{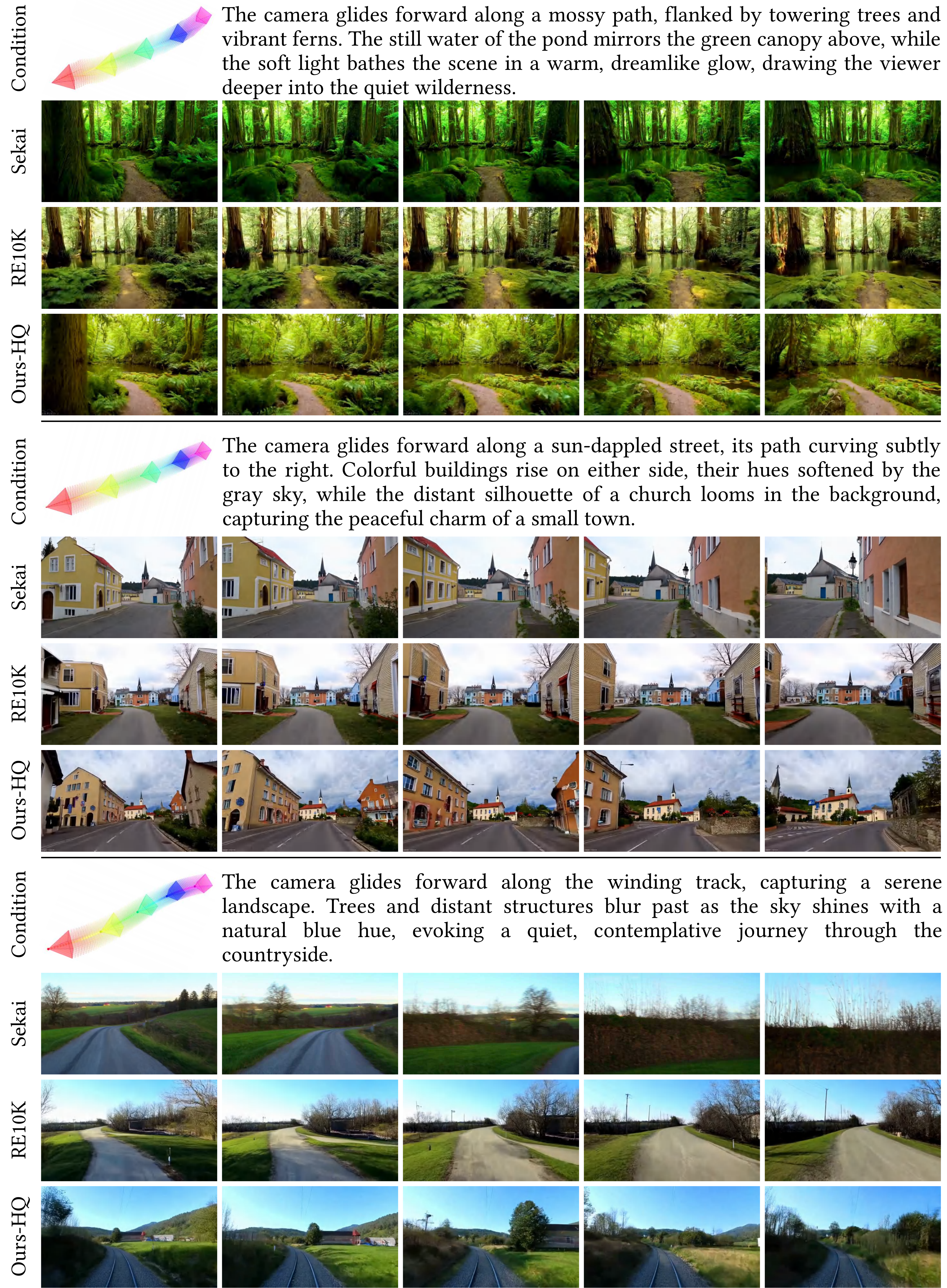}
    \end{overpic}
    \caption{
    \textbf{Different training datasets performance on SpatialVID samples.}
    }
    \label{fig:cvd_qualitative_supp_svid}
\end{figure*}

% qualitative
\begin{figure*}
    \centering
    \vspace{-4mm}
    \begin{overpic}[width=.9\textwidth,
    % \begin{overpic}[width=1.\columnwidth,
    clip
    ]{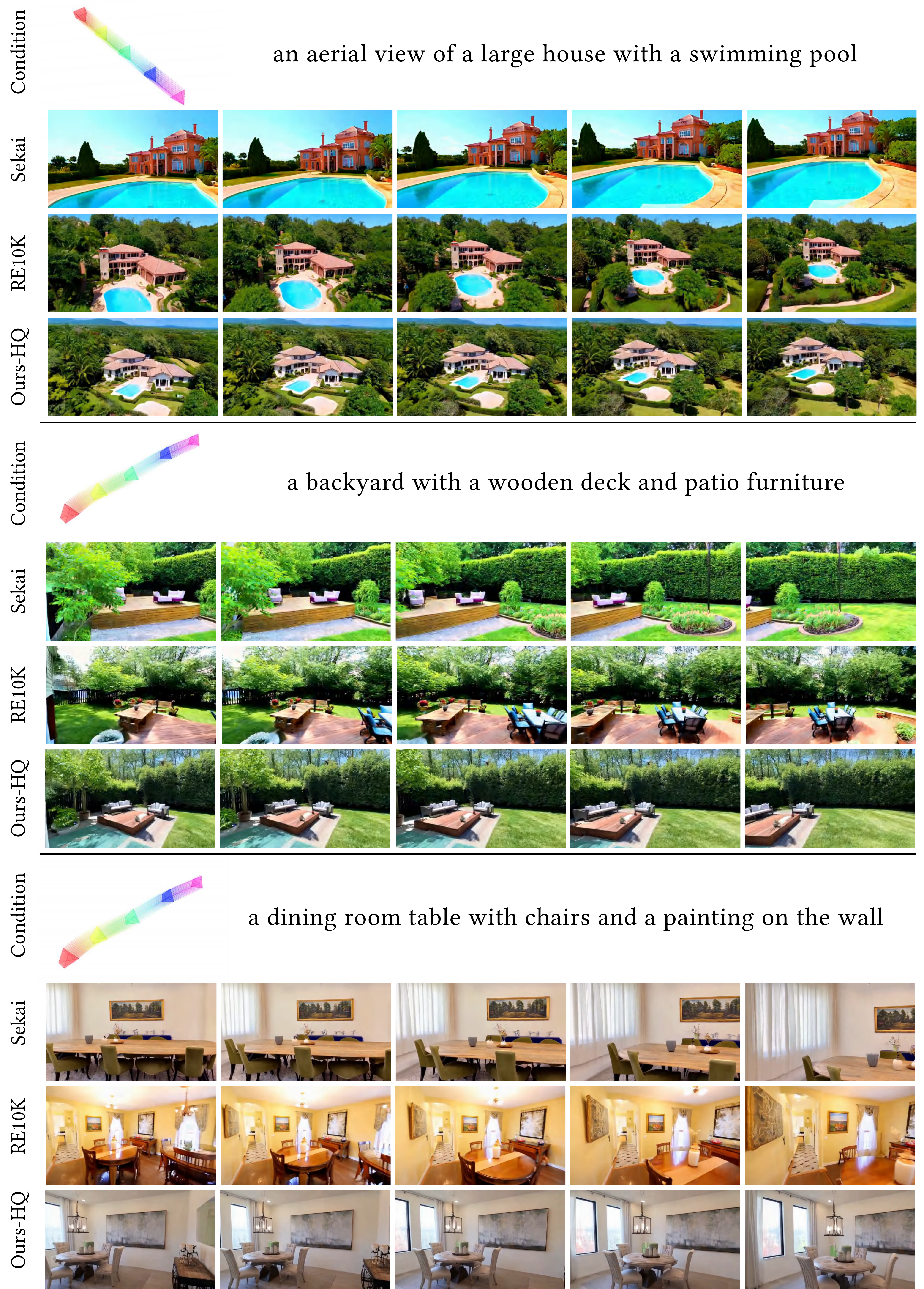}
    \end{overpic}
    \caption{
    \textbf{Different training datasets performance on RealEstate10K samples.}
    }
    \label{fig:cvd_qualitative_supp_r10k}
\end{figure*}

% qualitative
\begin{figure*}
    \centering
    \vspace{-4mm}
    \begin{overpic}[width=.9\textwidth,
    % \begin{overpic}[width=1.\columnwidth,
    clip
    ]{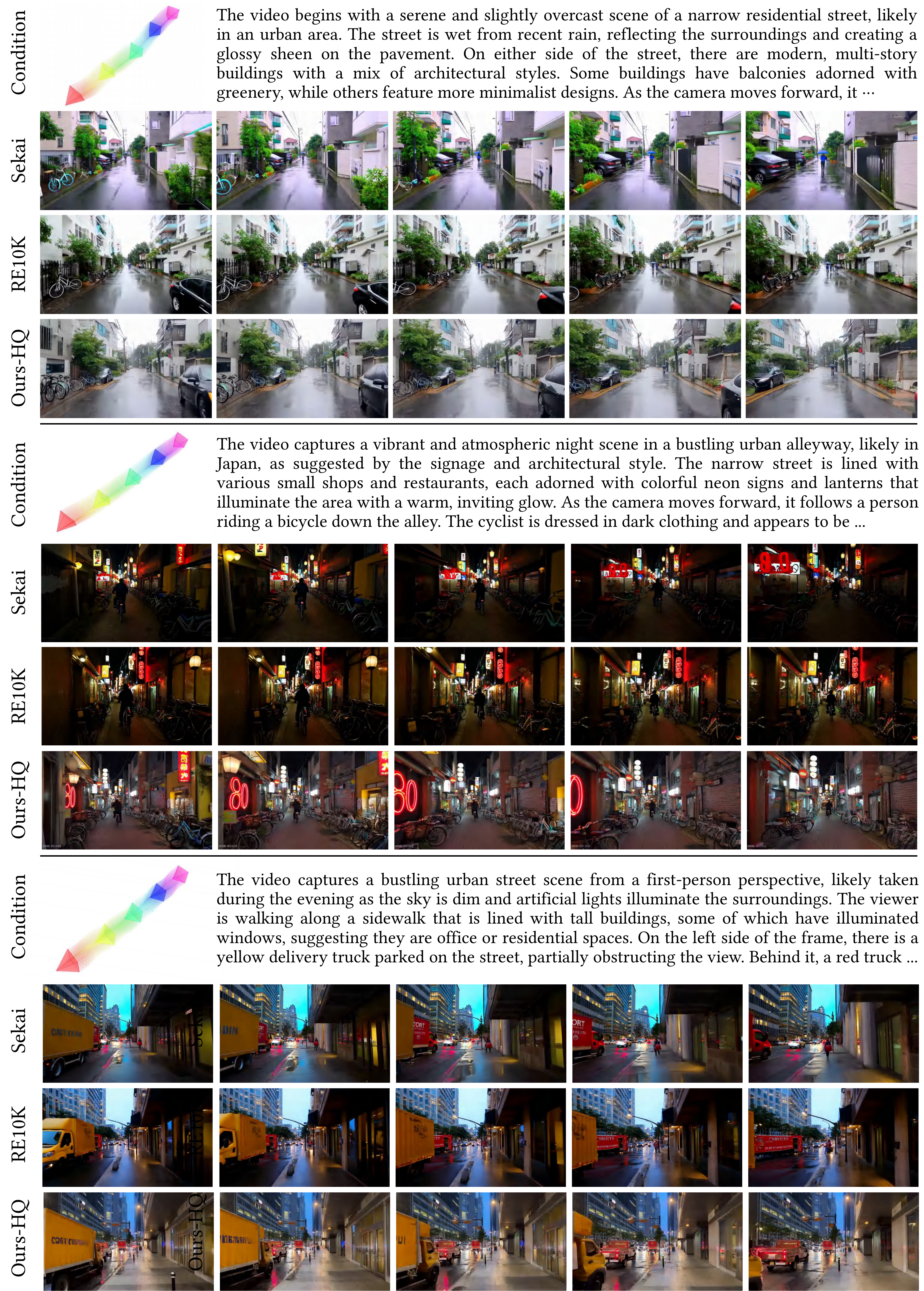}
    \end{overpic}
    \caption{
    \textbf{Different training datasets performance on Sekai-Real samples.}
    }
    \label{fig:cvd_qualitative_supp_sekai}
\end{figure*}

\begin{figure*}% [h!]
    % \vspace{-18mm}
    \centering
    \begin{overpic}[width=.95\textwidth,
    % \begin{overpic}[width=1.\columnwidth,
    % height=6cm,
    % trim=40 140 20 90,
    clip
    ]{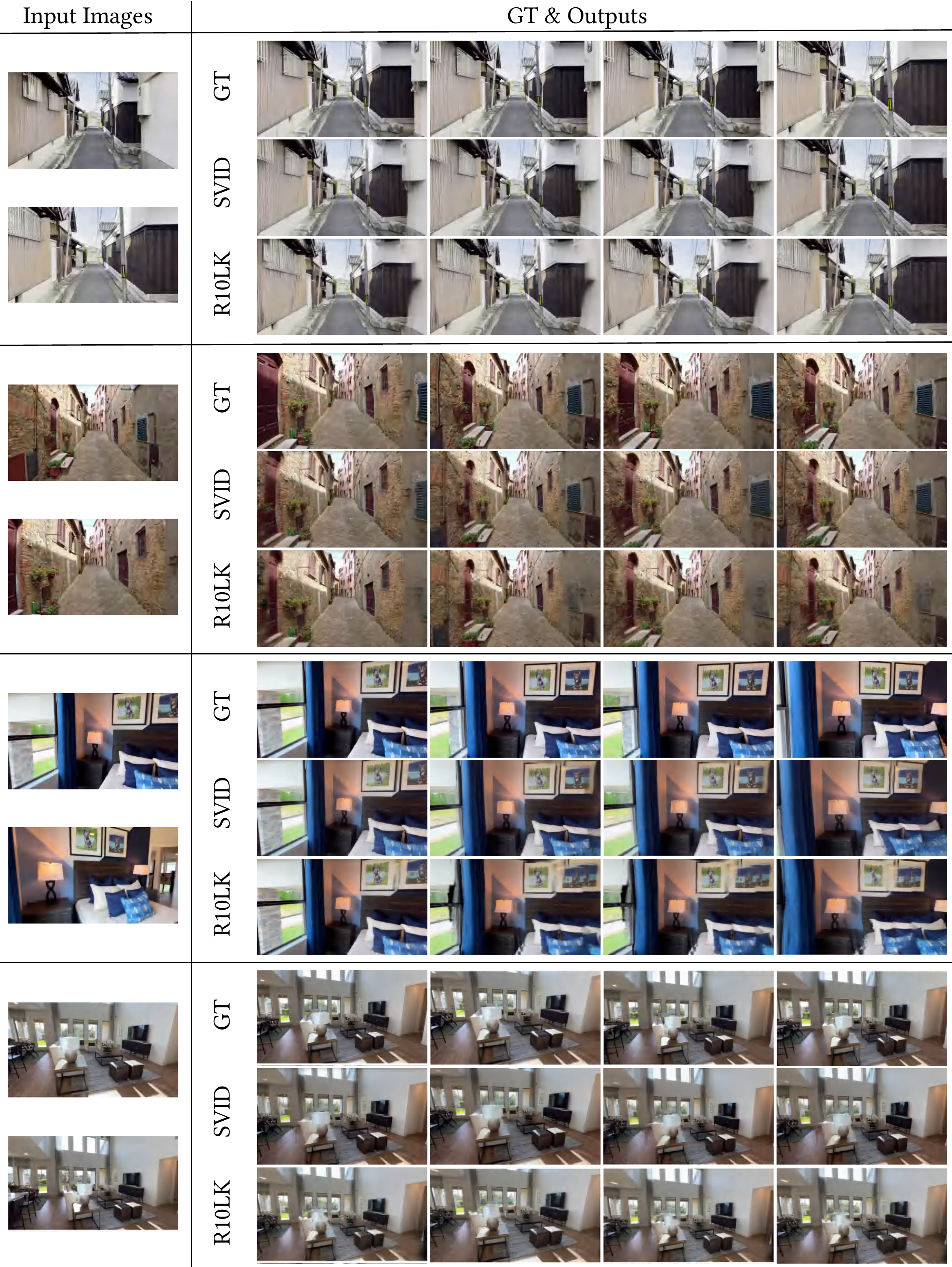}
    % \vspace{-8mm}
    \end{overpic}
    \caption{\textbf{GS-LRM qualitive comparison on SpatialVID}.}

        \label{fig:lrm_comp}
    % \vspace{-6mm}
\end{figure*}

\section{Validation Tasks}
\label{sec:supp:validation}

This section presents the implementation details and qualitative results of validation tasks conducted using the \DATAname~dataset. We focus on three representative paradigms, camera-controlled video generation, spatial reconstruction, and view-consistent rendering, to demonstrate the utility and quality of SpatialVID annotations.

\subsection{Implementation Details}

% implementation 范式：
% \wjh{template}
% All experiments are conducted on [Dataset1]~\citep{ref}, [Dataset2]~\citep{ref}, and our \DATAname-HQ~dataset. For fair comparison, we follow the same data split and preprocessing protocol as [Baseline/Reference]. All models are initialized from the publicly released [ModelName] checkpoint pretrained on [Dataset], ensuring consistent architecture and capacity. During training, input frames are resized to $[H]\times[W]$ and randomly augmented with [augmentations, e.g., horizontal flip or color jitter]. We train for [XXK] steps (or [XX] epochs) with a global batch size of [XX], using the [OptimizerName] optimizer with an initial learning rate of $[lr]$ and a [scheduler type, e.g., cosine decay or linear warm-up]. Unless otherwise specified, we adopt [loss terms, e.g., $\mathcal{L}_{\text{recon}}$, $\mathcal{L}_{\text{motion}}$, $\mathcal{L}_{\text{cam}}$], combined as $\mathcal{L}_{\text{total}} = \lambda_1 \mathcal{L}_{\text{recon}} + \lambda_2 \mathcal{L}_{\text{motion}} + \lambda_3 \mathcal{L}_{\text{cam}}$. All experiments are run on [\#GPUs] $\times$ [GPU model] using mixed precision with [framework name and version]. 

\noindent \textbf{Camera-Controlled Video Generation}.
Experiments are conducted on Sekai-Real~\citep{li2025sekai}, RealEstate10K~\citep{zhou2018stereo}, and our \DATAname-HQ~dataset. Since RealEstate10K does not include textual annotations, we adopt the text captions provided by CameraCtrl~\citep{he2024cameractrl}. For fair comparison, we follow the original train/test split for RealEstate10K, randomly sample 10K training clips from Sekai-Real, and use all high-quality clips from \DATAname-HQ. For \DATAname-HQ, the training captions are derived from the \textit{Immersive Shot Summary} component of our structured annotations. The DiT-based models are initialized from the publicly released TI2V-Wan2.2-5B checkpoint~\citep{wan2025wan}, ensuring consistent architecture and capacity across datasets. During training, the camera encoder, projector, and self-attention modules are learnable, while all remaining components are frozen. LInput frames are resized to $382\times480$ with a sequence length of 81 frames. We train for 20K steps using a global batch size of 32, the AdamW optimizer with an initial learning rate of $1\times10^{-5}$, a cosine decay schedule, and a warm-up period of 2K steps. Unless otherwise specified, camera conditioning follows the injection scheme introduced in ReCamMaster~\citep{bai2025recammaster}. For each frame, the $3{\times}4$ camera extrinsic matrix (12 parameters) is passed through a learnable linear layer $\text{cam\_encoder}\in\mathbb{R}^{12\times d}$ to project it into the feature dimension $d$. The resulting embedding is combined with visual tokens through a lightweight per-block \textit{projector} ($\mathbb{R}^{d\times d}$) initialized as an identity mapping to preserve the pretrained feature scale.

\noindent \textbf{CUT3R Fine-tuning}.
We follow the official fine-tuning protocol of CUT3R~\citep{cut3r} and initialize from the publicly released \texttt{cut3r\_512\_dpt\_4\_64} checkpoint. Training is performed with a global batch size of 64 using the AdamW optimizer with a learning rate of $1\times10^{-6}$, a weight decay of 0.05, and a total of 6,500 iterations. We fine-tune on long video sequences ranging from 4 to 64 frames, with each frame resized such that the longer side does not exceed 512 pixels. During training, the encoder is kept frozen while the decoder and output heads are updated to better align CUT3R’s geometric predictions with the spatially consistent captions and camera poses provided by \DATAname.

\noindent \textbf{VGGT Fine-tuning}.
We fine-tune VGGT~\citep{wang2025vggt} on SpatialVID-HQ together with most of its original training data. For fair comparison, an additional model is fine-tuned using the same original data without SpatialVID-HQ. We initialize from the publicly released VGGT checkpoint and follow the original training strategy, keeping the DINO backbone frozen to ensure consistent architecture and capacity. Training is performed for 40K steps, using the AdamW optimizer with an initial learning rate of $2\times10^{-4}$. Unless otherwise specified, all remaining hyperparameters follow the default VGGT configuration.

\noindent \textbf{Large Reconstruction Models}.
% wjh revise 1110
All experiments are conducted on RealEstate10K~\citep{zhou2018stereo} and our \DATAname-HQ~dataset. For fair comparison, both datasets are trained with an equal amount of 60K video clips. In each epoch, random image pairs are sampled, using two views as input and four intermediate views as supervision. All models are trained from scratch to ensure consistent architecture and capacity. Following the GS-LRM~\citep{zhang2024gs} training protocol, we adopt a two-stage schedule: the first stage trains at a resolution of $180\times320$ for 15K steps, followed by a high-resolution stage at $360\times640$ for an additional 45K steps, totaling 60K steps. Training is performed with a global batch size of 32 using the AdamW optimizer, an initial learning rate of $2\times10^{-5}$, a cosine decay schedule, and 2K warm-up steps. Unless otherwise specified, the total loss combines pixel-level, perceptual, and depth-smoothness objectives: $\mathcal{L}_{\text{total}} = \lambda_1 \mathcal{L}_{\text{mse}} + \lambda_2 \mathcal{L}_{\text{lpips}} + \lambda_3 \mathcal{L}_{\text{reg}}$, where $\lambda_1=1.0$, $\lambda_2=0.5$, and $\lambda_3=0.25$. The regularization term $\mathcal{L}_{\text{reg}}$ encourages depth smoothness by penalizing abrupt depth discontinuities.

\subsection{Qualitative Results}

We provide additional qualitative comparisons of Camera-controlled video generation and Novel View Synthesis.
% The camera-controlled video generation results (\figref{fig:cvd_qualitative_supp_svid}, \figref{fig:cvd_qualitative_supp_r10k}, \figref{fig:cvd_qualitative_supp_sekai}) demonstrate precise trajectory adherence under complex camera motions, with realistic spatial continuity and dynamic visual consistency.
The camera-controlled video generation results (\figref{fig:cvd_qualitative_supp_svid}, \figref{fig:cvd_qualitative_supp_r10k}, \figref{fig:cvd_qualitative_supp_sekai}) show that the model trained on \textit{SpatialVID-HQ} precisely follows complex camera trajectories while maintaining realistic spatial continuity and dynamic visual coherence. Moreover, the model demonstrates improved prompt understanding, enabling the generation of more accurate and visually convincing environmental details such as trees and decorations.
The novel view synthesis results (~\figref{fig:lrm_comp}) highlight how SpatialVID supports robust geometry learning, maintaining consistent spatial layouts and detailed texture synthesis across diverse motion trajectories. 

By integrating explicit 3D motion control with rich textual semantics, \DATAname endows physically grounded video generation, dynamic scene synthesis, and spatial intelligence tasks with robust 3D inductive biases. Comprehensive experimental results validate SpatialVID’s effectiveness across diverse tasks, laying a solid foundation for research in the field of spatial intelligence.